%% file: main.tex
\documentclass{article}
\pdfoutput=1
\usepackage[table]{xcolor}
\usepackage{nips15submit_e,times}

\usepackage{comment}
\usepackage{hyperref}
\usepackage{url}
\usepackage{natbib}
\usepackage{amsthm}
\usepackage{algorithm}
\usepackage{algpseudocode}
\usepackage{longtable}
\usepackage{afterpage}
\usepackage{amsmath}
\usepackage{amssymb}
\usepackage{graphicx}
\usepackage{mathtools}
\DeclareMathOperator*{\argmin}{argmin}
\DeclareMathOperator*{\argmax}{argmax}
\DeclarePairedDelimiter\brangle{\langle}{\rangle}

\newtheorem{definition}{Definition}

\newtheorem{theorem}{Theorem}

\newcommand{\y}{{\bf y}} % assignment to all latent variables
\newcommand{\Y}{\mathcal{Y}} % set of all latent variables
\newcommand{\m}{m} % number of latent variables
\newcommand{\n}{n} % number of observed variables

\title{Anchored Discrete Factor Analysis}

\definecolor{lightgray}{gray}{0.9}

\let\oldtabular\tabular
\let\endoldtabular\endtabular
\renewenvironment{tabular}{\rowcolors{2}{lightgray}{white}\oldtabular}{\endoldtabular}

\let\oldlongtable\longtable
\let\endoldlongtable\endlongtable
\renewenvironment{longtable}{\rowcolors{2}{white}{lightgray}\oldlongtable} {
\endoldlongtable}

\LTcapwidth=\textwidth

\nipsfinalcopy % Uncomment for camera-ready version

\title{Anchored Discrete Factor Analysis}
\author{Yoni Halpern$^1$, Steve Horng$^2$, David Sontag$^1$ \\
$^1$New York University, $^2$Beth Israel Deaconess Medical Center}
\begin{document}
\maketitle
\begin{abstract}
\input{tex/abstract}
\end{abstract}
\input{tex/intro}

\input{tex/background}
\input{tex/recovering_moments}
\input{tex/learningY}

\input{tex/factor_loadings}
\input{tex/experiments}

\input{tex/Discussion}
\input{tex/Acknowledgments.tex}
\bibliographystyle{authordate1}
\small{
\bibliography{nips14}
}
\appendix
\input{tex/supplemental_materials}
\end{document}

%% file: tex/abstract.tex
%auto-ignore
%!TEX root = ../main.tex
We present a semi-supervised learning algorithm for learning discrete factor analysis models with arbitrary structure on the latent variables. Our algorithm assumes that every latent variable has an ``anchor'', an observed variable with only that latent variable as its parent. Given such anchors, we show that it is possible to consistently recover moments of the latent variables and use these moments to learn complete models. We also introduce a new technique for improving the robustness of method-of-moment algorithms by optimizing over the marginal polytope or its relaxations. We evaluate our algorithm using two real-world tasks, tag prediction on questions from the Stack Overflow website and medical diagnosis in an emergency department.

% We apply the new learning algorithm to two real world datasets and show that it learns meaningful representations. 

%% file: tex/intro.tex
%auto-ignore
%!TEX root = ../main.tex

\section{Introduction}
\label{sec:intro}

Estimating the parameters and structure of Bayesian networks from incomplete data is a fundamental task in many of the social and natural sciences. 
We consider the setting of {\em latent variable} models, where certain variables of interest are never directly observed, but their effects are discernible in the interactions of other directly observable variables. 
For example, in the clinical setting, diseases themselves are rarely observable, but their presence and absence are inferred from the combined evidence of patient narratives, lab tests, and other measurements.
A full understanding of the relationships between diseases, how they interact with each other and how they present in individuals is still largely unknown. % and is the ultimate goal of personalized medicine. 
Another example is from the social sciences, where we may wish to know how opinions and beliefs interact or change over time. 
Surveys, textual analysis, and actions (e.g., voting registration, campaign donations) provide a noisy and incomplete view of people's true beliefs, and we would like to build models to describe and analyze the underlying world-views.

Estimating models involving latent variables is challenging for two reasons.
First, without some constraints on the model, the model may be {\em non-identifiable}, meaning that there may exist multiple parameter settings that cannot be distinguished based on the observed data. This non-identifiability severely diminishes the interpretability of the learned models, and has been the subject of many critiques of factor analysis and similar techniques. 
Second, even in the identifiable setting, finding the model that best describes the data is often computationally intractable, even for seemingly simple models. 
For example, without making assumptions on the data generation process, finding the maximum likelihood latent Dirichlet allocation model is NP-hard \citep{AGM}. %arora2005learning, AGKM}.
% TODO: Arora & Kannan don't prove mixture of Gaussians in NP hard... find better reference.

We present a new efficient algorithm for discrete factor analysis, learning models involving binary latent variables and non-linear relationships to binary observed variables \citep{martin1995discrete,Singliar, wood06, JerHalSon_nips13}. 
Our algorithm is semi-supervised, requiring that a domain expert identify {\em anchors}, observed variables that are constrained to have only a single parent among the latent variables. Given these anchors, we show how to learn factor analysis models with arbitrary structure in the latent variables. The anchor conditions we present are closely related to the {\em exclusive views} conditions used by \citet{chaganty2014graphical}, which enable learning model parameters provided that each clique has a set of observations that satisfy certain independence criteria. However, identifying exclusive views requires that the structure of the latent variable model be known in advance. This is a significant shortcoming in real-world settings, where the network structure is almost never known and its recovery is the main goal of data analysis. 

\begin{figure}[t]
    \begin{center}
        \includegraphics[scale=0.4]{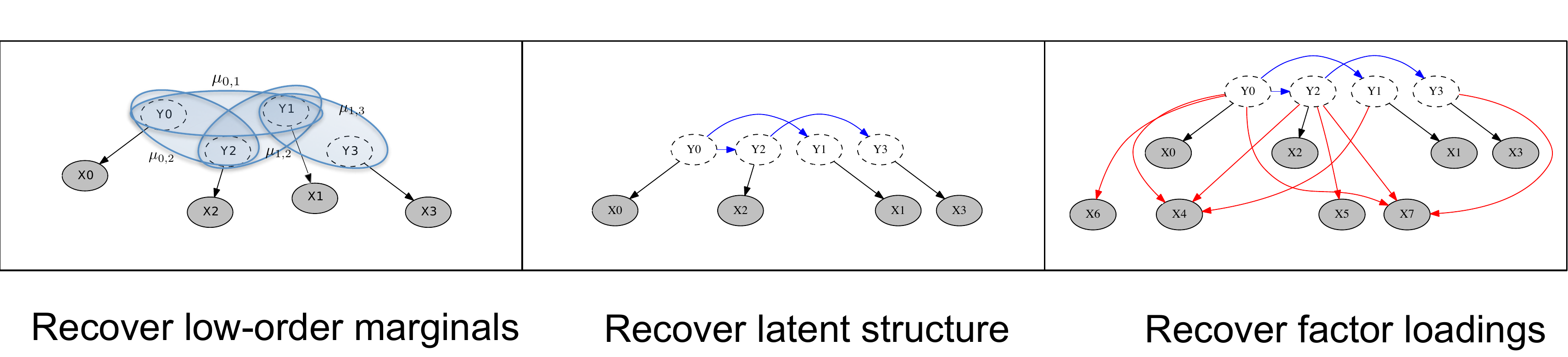}
    \end{center}
\vspace{-2mm}
    \caption{\small Illustration of our algorithm for anchored discrete factor analysis. 
%    We take as input a simple model where only the conditional probabilities of the anchors given the latent variables are known.
   First, we learn the low-order moments of the latent variables by solving an optimization problem involving moments of the observed anchors. Second, we use these moments to learn a Bayesian network describing the joint distribution of the latent variables. Finally, we learn the conditional distributions for all non-anchor observations.
    \label{fig:algorithm}}
\end{figure}

Our learning algorithm is based on the method-of-moments,
which over the last several years has led to polynomial-time algorithms for provably learning a wide range of latent variable models \citep[e.g., ][]{animaLatentTrees, AHK, AroraEtAl_icml13}. 
The algorithm, summarized in Algorithm~\ref{alg:overall} and illustrated in Figure~\ref{fig:algorithm}, proceeds in steps, first learning the structure on the latent variables, and then learns edges from the latent to the observed variables.

{\bf Contributions:} \quad The contributions of this paper are as follows:
\begin{itemize}
\item We define anchored factor analysis, a subclass of factor analysis models, where the structure of the latent variables can be recovered from data. 
\item We derive a method-of-moments algorithm for recovering both the latent structure and the factor loadings, including an efficient method for the case of tree-structured latent variables.
\item We show that moment recovery can be improved by using constrained optimization with increasingly tight outer bounds on the marginal polytope, suggesting a method of increasing robustness of method-of-moments approaches in the presence of model misspecification.
\item We present experimental results, learning interpretable factor analysis models on two real-world multi-label text corpora, outperforming discriminative baselines.
\end{itemize}

% In this paper we consider discrete factor analysis \citep{martin1995discrete,Singliar, wood06, JerHalSon_nips13}. Compared to factor analysis, which assumes a Gaussian distribution over the latent variables and a linear observation model, the models that we consider allow for arbitrary distributions on discrete latent variables and non-linear relationships with discrete observations. We consider a semi-supervised learning setting, similar to confirmatory factor analysis \citep{joreskog69}, where a domain expert identifies the latent variables to be modeled by providing a small number of structural constraints. Such settings may be particularly useful in uncovering interpretable latent variables in the medical domain where latent variables of interest such as disease states are known to be highly interacting and often unknown to some degree. 

% Even in models that are identifiable, computational and statistical efficiency are important to consider as well. 
% Maximum likelihood estimation of the parameters in these models that contain latent variables is a difficult task since the likelihood objective is non-concave. 
\begin{algorithm}[t]
\caption{\label{alg:overall} High level algorithm for anchored factor analysis model learning.}
\begin{algorithmic}[1]
\Require{Dataset $\mathcal{D}$, Anchors $\mathcal{A}$}
\State $\mu$ = RecoverMarginals($\mathcal{D}, \mathcal{A}, P(A|Y)$) (Section~\ref{sec:recovering})
\State $P(\mathcal{Y})$ = BuildLatentModel($\mu$) (Section~\ref{sec:structure})
\State $P(\mathcal{X|Y})$ = LearnFactorLoadings($\mu, P(\mathcal{Y})$) (Section~\ref{sec:loadings})
\State $P(\mathcal{X,Y}) = P(\mathcal{Y})P(\mathcal{X|Y})$
\Ensure{ADFA model: $P(\mathcal{X,Y})$}
\end{algorithmic}
\end{algorithm}

{\bf Road map:} \quad The paper proceeds according to this high level structure:
\begin{itemize}
\item Section~\ref{sec:notation} formally defines anchor variables.
\item Section~\ref{sec:recovering} reviews how anchor variables can be used to recover moments of the latent variables.
\item Section~\ref{sec:learning} describes a family of latent variable models that can be learned using the recovered latent moments, and outlines the learning procedure, which consists of two parts:
learning a structure for the latent variables and
learning connections between the latent variables and observed variables.
\item Section~\ref{sec:experiments} compares the models learned using the method described here to other standard baselines on real-world prediction tasks.
\end{itemize}

%% file: tex/background.tex
%auto-ignore
%!TEX root = ../main.tex

\section{The anchored factor analysis task}
\label{sec:notation}

{\bf Notation:} We use uppercase letters to refer to random variables (e.g., $Y_i, X_j$) and corresponding lowercase letters to refer to the values. For compactness, we use superscript parenthetical values to denote a variable taking a particular value (e.g., $P(x_j^{(0)})$ is equivalent to $P(X_j = 0)$).

{\bf Anchored factor analysis:} In this paper, we will discuss learning binary latent variable models. We divide the variables of our model into two classes: {\em observed} variables $\mathcal{X} = \{X_{1}, ..., X_{\n}\}$ and unobserved or {\em latent} variables $\Y = \{Y_1, ..., Y_\m\}$, where all variables have binary states $0$ or $1$.

As in other factor analysis models, we assume that all dependencies between the observations are explained by the latent variables. This implies that the observed distribution has the following factorized form: $P(X_1, ..., X_\n) = \sum_{y}P(y)\prod_{j=1}^n P(x_j \mid y)$. 

%We parameterize $P(\y)$ as a Bayesian network on a graph $\mathcal{G}$, i.e. $P(y_1, ..., y_m) = \prod_{i=1}^\m P(y_i \mid \y_{Pa(i)})$ where $Pa(i)$ are the parents of $i$ according to $\mathcal{G}$. 
%Our algorithms for learning the factor loadings $P(x_j \mid \y)$ assume that these are parameterized as {\em noisy-or} distributions (see Section~\ref{sec:loadings}). 
We focus on a particular class of latent variable models, that we term {\em anchored} models. 
A latent variable model is anchored if for every $Y_i \in \mathcal{Y}$, there is a corresponding observed variable $A_{Y_i} \in \mathcal{X}$ that fits the following definition:
\begin{definition}
\label{def:anchor}
{\bf Anchor variables}: An observed variable $X_j \in \mathcal{X}$ is called an {\em anchor} for a latent variable $Y_i \in \mathcal{Y}$ if 
$X_j \not \perp Y_i$ and  
$X_j \perp Y_k | Y_i $ for all other $Y_{k \neq i} \in \mathcal{Y}$. 
\end{definition}
Using the language of directed probabilistic graphical models, we say that $X_j$ is an anchor of $Y_i$ if $Y_i$ is the sole parent of $X_j$. We use the notation $A_{Y_i}$ to refer to the anchor of $Y_i$ or $A_{\mathcal{Z}}$ to refer to the set of anchors of a set of latent variables $\mathcal{Z} \subseteq \mathcal{Y}$. 

In this work, we assume that anchors are identified as an input to the learning algorithm. Specifying anchors can be a way for experts to inject domain knowledge into a learning task. In section~\ref{sec:experiments} we show that even anchors obtained from simple rules can perform well in modeling. 

% \subsection{Anchors as noisy labels}
% Anchors are related to noisy labels in a multi-label learning setting. 
% For discrete variables, the definition of anchors above is a multi-label extension of the {\em class-dependent noise} setting \citep[e.g.][]{natarajan2013learning,ScottBH13}. 
% Learning with noisy labels has been studied extensively, and the sensitivity of different classifiers have been studied under various noise models. 
% This connection to noisy labels informs the baseline approaches in our experimental results. 

%% file: tex/recovering_moments.tex
%auto-ignore
%!TEX root = ../main.tex

\section{Recovering moments of latent variables}
\label{sec:recovering}

\subsection{Previous work -- exclusive views}
\label{sec:exclusiveviews}
For the purpose of exposition, we assume that every anchor has a conditional distribution (i.e., $P(A_i | Y_i)$) which can be estimated consistently, and leave the discussion of estimating these conditional distributions from data to a later section (Section~\ref{sec:noise_rates}).

For a set of latent variables $\mathcal{Z} \subseteq \mathcal{Y}$, 
\citet{chaganty2014graphical} show how to recover the marginal probabilities $P(\mathcal{Z})$ using only the observed anchors, $\mathcal{A}_\mathcal{Z}.$ 
Eq.~\ref{eq:anchor_moments} relates the observed distribution $P(\mathcal{A}_{\mathcal{Z}})$ to the unobserved moments $P(\mathcal{Z})$:
\begin{equation}
\label{eq:anchor_moments}
P(\mathcal{A}_\mathcal{Z}) = \sum_{z}P(z)P(\mathcal{A}_\mathcal{Z}|z) = \sum_{z}P(z)\prod_{i=1}^{|\mathcal{Z}|} P(A_i|z_i).
\end{equation}
The first equality simply comes from marginalizing the latent variables $\mathcal{Z}$. 
The second uses the assumption that an anchor is independent of all other variables conditioned on its parent. 
Since we assume that $P(A_i|Z_i)$ can be estimated, Eq.~\ref{eq:anchor_moments} gives $2^{|\mathcal{Z}|}$ linear equations for $2^{|\mathcal{Z}|}$ unknowns. 
Since the linear equations can be shown to be independent (see supplementary materials), the distribution of the latent variables $P(\mathcal{Z})$ can be estimated from the observed distribution $P(\mathcal{A}_{\mathcal{Z}})$. 
This can be trivially extended to a case where $\mathcal{Z}$ includes both latent and observed variables by noticing that every observed variable can act as its own anchor.

Let $R_{\mathcal{Z}}$ be the Kroneker product of the conditional anchor distributions, $R_{\mathcal{Z}} = \otimes_{k=1}^{|\mathcal{Z}|} P(A_k | Z_k)$,
and $\mu_{\mathcal{Z}}$ be a vector of dimension $2^{|\mathcal{Z}|}$ of probabilities for $P(\mathcal{Z})$.
To recover $P(\mathcal{Z})$, we seek to find a distribution that minimizes the divergence between the expected marginal vector of the anchors, $R_{\mathcal{Z}} \mu_{\mathcal{Z}}$, and the observed marginal vector of the anchors, $\mu_{\mathcal{A}_{\mathcal{Z}}}$. We solve this as a constrained optimization problem, where $D$ is any Bregman divergence and $\Delta$ is the probability simplex:
\begin{equation}
\label{eq:anchor_opt}
\hat{\mu}_{\mathcal{Z}}  =  \argmin_{\mu \in \Delta} D\left( \mu_{\mathcal{A}_{\mathcal{Z}}}, R_{\mathcal{Z}} \mu\right).
\end{equation} 
%\end{equation}
%In section~\ref{sec:robust} we will show that constraining this minimization problem yields more robust methods recovering the moments. The estimator described in Equation~\ref{eq:moment_estimator} and the robust versions described in Section~\ref{sec:robust} 
\cite{chaganty2014graphical} show the consistency of this estimator using both L2 distance and KL divergence for $D$. The consistency of the estimator means that the recovered marginals $\hat{\mu}_\mathcal{Z}$ will converge to the true values, $\mu_\mathcal{Z}$, assuming that the anchor assumption is correct. %Using these estimators, we can proceed as though we have fully observed moments on $\mathcal{Z}$.

%\section{Algorithmic Improvements}
\subsection{Robust moment recovery -- connection to variational inference}
\label{sec:robust}
%When we only have estimates of $\hat{P}(\mathcal{A})$ and $\hat{P}(A_i|Z_i)$, the system of equations in Eq.~\ref{eq:anchor_matrix} are unlikely to  hold perfectly.

% TODO: cite this somewhere. Perhaps in discussion is most appropriate.
%Similar approaches were developed in statistics to improve the robustness of estimation of factor analysis from discrete data \citep{knol89}. 

Our algorithm's theoretical guarantees, as with other method-of-moments approaches, assume that the data is drawn from a distribution in the model family. In our setting, this means that the anchors must perfectly satisfy Definition~\ref{def:anchor}. We introduce a new approach to improve the robustness of method-of-moment algorithms to model misspecification. Our approach can be used as a drop-in replacement for parameter recovery in the exclusive views framework of \citet{chaganty2014graphical}. For simplicity, we describe it here in the context of anchors.

Consider, for example, the case of learning a tree-structured distribution on the latent variables. Before running the Chow-Liu algorithm, we would need to estimate the edge marginals $\mu_{ij}(y_i, y_j)$ for every pair of random variables $Y_i, Y_j$. \citet{chaganty2014graphical}'s approach is to solve $n \choose 2$ independent optimization problems of the form given in Eq.~\ref{eq:anchor_opt}, resulting in a set of estimates $\hat{\mu}_{ij}(y_i, y_j)$. Our key insight is that since the true edge marginals $\{\mu_{ij}(y_i, y_j) : i,j\in \mathcal{Y}\}$ all derive from $P(\mathcal{Y})$, i.e. $\mu_{ij}(y_i, y_j) = \sum_{\y_{\backslash i,j}} P(\y)$, there are additional constraints that they must satisfy. For example, the true edge marginals must satisfy the {\em local consistency} constraints:
\begin{equation}
\sum_{y_i} \mu_{ij}(y_i, y_j) = \sum_{y_k} \mu_{jk}(y_j, y_k) \quad \forall i, j, k\in \mathcal{Y} \textrm{ and } y_j.
\label{eq:local_consistency}
\end{equation}
%
%In this section, we introduce a new approach for improving the robustness of method-of-moment algorithms based on optimization over (relaxations of) the {\em marginal polytope}. 
%Rather than requiring equality, we attempt to minimize a divergence between the right and left sides of Equation~\ref{eq:anchor_matrix} to obtain an estimate of the desired marginal vector.
More generally, $\mu$ must lie in the {\em marginal polytope}, $\mathcal{M}$, consisting of the space of all possible marginal distribution vectors that can arise from any distribution \citep{WJ}. Note that there exist vectors $\mu$ which satisfy the local consistency constraints but are not in the marginal polytope. % (also non-negativity and normalization)
%\begin{equation*}
%\mathcal{M} = \left\{ \mu \in \mathbb{R}^d | \exists p \hspace{1ex} s.t. \hspace{1ex} \text{marginals of } p = \mu \right\}. 
%\end{equation*}
%
The marginal polytope has been widely studied in the context of probabilistic inference in graphical models. MAP inference corresponds to optimizing a linear objective over the marginal polytope, and computing the log-partition function can be shown to be equivalent to optimizing a non-linear objective over the marginal polytope \citep{WJ}.
%Approximate inference algorithms circumvent the complexity of the marginal polytope by instead optimizing over an outer bound given by a small number of constraints.
Optimizing over the marginal polytope is NP-hard, and so approximate inference algorithms such as belief propagation instead optimize over the outer bound given by the local consistency constraints. There has been a substantial amount of work on characterizing tighter relaxations of the marginal polytope, all of which immediately applies to our setting \citep[e.g., ][]{SonJaa_nips07}. %, Peng12}.

%We minimize the Kullback-Leibler divergence, which previous work has shown to have improved asymptotic efficiency compared to other divergence measures \citep{chaganty2014graphical}.
Putting these together, we obtain the following optimization problem for robust recovery of the true moments of the latent variables from noisy observations of the anchors:
\begin{equation}
\label{eq:anchor_opt_matrix}
\hat{\mu} = \argmin_{\mu \in \mathcal{P}} \sum_{\mathcal{Z} \subseteq \mathcal{Y} : |\mathcal{Z}| \leq K} D_{KL} \left(\mu_{\mathcal{A}_{\mathcal{Z}}}, R_{\mathcal{Z}} \mu_{\mathcal{Z}} \right),
\end{equation}
where $K$ is the size of the moments needed within the structure learning algorithm (e.g., $K=2$ for a tree-structured distribution) and $\mathcal{P}$ denotes a relaxation of the marginal polytope.

We use the conditional gradient, or Frank-Wolfe, method to minimize \eqref{eq:anchor_opt_matrix} \citep{Frank-Wolfe}. %, described in Algorithm~\ref{alg:FW} for optimizing concave functions over a compact convex set.
Frank-Wolfe solves this convex optimization problem by repeatedly solving linear programs over $\mathcal{P}$. When $\mathcal{P}$ corresponds to the local consistency constraints or the cycle relaxation \citep{SonJaa_nips07}, these linear programs can be solved easily using off-the-shelf LP solvers. Alternatively, when there are sufficiently few variables, one can optimize over the marginal polytope itself. For this, we use the observation of \citet{belangerWorkshop2013} that optimizing a linear function over the marginal polytope can be performed by solving an integer linear program with local consistency constraints. 
In the experimental section we show that constrained optimization improves the robustness of the moment-recovery step compared to unconstrained optimization and that using increasingly tight approximations of the marginal polytope within the conditional gradient procedure yields increasingly improved results.

Constrained optimization has been used previously to improve the robustness of method-of-moments results~\citep{exteriorpoint}. 
Our work differs in that the constrained space naturally coincides with the marginal polytope, which allows us to leverage the Frank-Wolfe algorithm for interior-point optimization and relaxations of the marginal polytope that have been studied in the context of variational inference. 

\subsection{Estimating anchor noise rates}
\label{sec:noise_rates}
 
Throughout we have assumed that the conditional probabilities of the anchors can be estimated from the observed data. This task of estimating label noise is the subject of a rich literature. In this section we describe four settings where that is a reasonable assumption. 

\begin{enumerate}
\item  {\bf Two or more anchors for $Y_i$ are provided:} If two anchors are provided, then the conditional distributions for both  can be estimated using a multi-view tensor decomposition method \citep{Berge,AHK}, where the third view is obtained using any other observed variable that is correlated with the anchors (see supplementary materials).

\item  {\bf Anchors are positive-only:} In the setting where anchors are positive-only (i.e. $P(Y_i | A_i)=1)$, we have the setting known as Positive and Unlabeled (PU) learning. In this setting, the noise rates of these noisy labels can be estimated using the predictions of a classifier trained to separate the positive from unlabeled cases \citep{elkan2008learning}.

\item  {\bf Mutually irreducible distributions:}  This setting, discussed in \cite{ScottBH13} can be described as requiring that in the data distribution there exists at least one unambiguous positive and negative case. Estimators for the noise rates under this setting are described in \citep{menon2015learning}.

\item  {\bf Some (partially) labeled data is available:}  In the medical diagnosis task that we consider in the experiments, we asked a physician a small number of questions about each patient as part of their regular workflow. This gave us {\em singly-labeled data}, where each data point observed $\mathcal{X}$ and $Y_i$ for a single $i$. Using a Chernoff bound, one can show that a small number of such observations suffices to accurately estimate an anchor's conditional distribution. 
\end{enumerate}

%\subsection{Reduction to fully observed setting}
%\label{sec:reduction}
%The anchor assumption, though seemingly mild in the setting where $n >> m$, is enough to allow for recovery of any marginal distribution of the latent and observed variables. 
%The key insight is that any marginal probability involving latent variables can be recovered using a marginal probability involving the appropriate anchors instead.
%

%% file: tex/learningY.tex
%auto-ignore
%!TEX root = ../main.tex
\section{Model learning}
\label{sec:learning}

\subsection{Learning $P(\mathcal{Y})$}
\label{sec:structure}

{\bf Structure learning background:} \quad Approaches for Bayesian network structure learning typically follow two basic strategies: they either search over structures $\mathcal{G}$ that maximize the likelihood of the observed data (score-based methods), or they test for conditional independencies and use these to constrain the space of possible structures. A popular scoring function is the BIC score \citep{Lam_MDL94, heckerman}:
\begin{equation}
\text{Score}_{BIC}(\mathcal{G}) =
\sum_{i=1}^\m N \hat{I}(Y_i ; Y_{Pa(i;\mathcal{G})}) - N \hat{H}(Y_i) - \log(N) 2^{|Pa(i; \mathcal{G})|}
\end{equation}
where $N$ is the number of samples and $\hat{I}, \hat{H}$ are the empirical mutual information and entropy respectively. $Pa(i; \mathcal{G})$ denotes the {\em parents} of node $i$, in graph $\mathcal{G}$. The last term is a complexity penalty that biases toward structures with fewer parameters. 
Once the optimal graph structure is determined, the conditional probabilities, $\theta$, that parametrize the Bayesian network are estimated from the empirical counts for a maximum likelihood estimate. 

BIC is known to be {\em consistent}, meaning that if the data is drawn from a distribution which has precisely the same conditional independencies as a graph $\mathcal{G}^*$, once there is sufficient data it will recover $\mathcal{G}^*$, i.e. $\mathcal{G}^* = \argmax_{\mathcal{G}} \text{Score}_{BIC}(\mathcal{G})$. 
In general, finding a maximum scoring Bayesian network structure is NP-hard \citep{Chickering96lns}. However, approaches such as integer linear programming \citep{jaakkola2010learning,gobnilp},  branch-and-bound \citep{Fan14tightening}, and greedy hill-climbing \citep{Teyssier+Koller:UAI05} can be remarkably effective at finding globally optimal or near-optimal solutions. 
Restricting the search to tree-structures, we can find the highest-scoring network efficiently using a maximum spanning tree algorithm \citep{chow1968approximating}. 

A different approach is to use conditional independence tests. Under the assumption that every variable has at most $k$ neighbors, these can be used to give polynomial time algorithms for structure learning \citep{PearlVerma,spirtes}, which are also provably consistent.

% TODO: this can be expanded a bit, e.g. by giving an algorithm box or more details... 

We utilize the fact that both score-based and conditional independence based structure learning algorithms can be run using derived statistics rather than the raw data. Suppose $P(\mathcal{Y})$ is a Bayesian network with maximum indegree of $k$. Then, to estimate the mutual information $\hat{I}$ and entropy $\hat{H}$ needed to evaluate the BIC score for all possible such graphs, we only need to estimate all moments of size at most $k+1$. For example, to search over tree-structured Bayesian networks, we would need to estimate $p(y_i, y_j)$ for every $i,j$.
Alternatively, we could use the polynomial-time algorithms based on conditional independence tests. These need to perform independence tests using conditioning sets with up to $k$ variables, requiring us to estimate all moments of order $k+2$.

We showed in Section~\ref{sec:recovering} how to recover these moments of the latent variables from moments of the anchor variables. Even though these latent variables are completely unobserved, we are able to use the recovered moments in structural learning algorithms designed for fully observed data. Together, this gives us a polynomial-time algorithm for learning the Bayesian network structure for the latent variables (Step 2 of Algorithm~\ref{alg:overall}) as stated in Theorem~\ref{thm:struct_consistency}. 

\begin{theorem}
\label{thm:struct_consistency} Let $\mathcal{G}(\mathcal{Y})$ be the graph structure of a Bayesian network describing the probability distribution, $P(\mathcal{Y})$, of the latent variables in an anchored discrete factor analysis model. Using the low-order moments recovered in Equation~\ref{eq:anchor_opt} in a structure learning algorithm which is consistent for fully-observed data, is a consistent structural estimator. That is, as the number of samples, $N \rightarrow \infty$, the structure recovered ,$\mathcal{G}^{\prime}(\mathcal{Y})$, is Markov equivalent to $\mathcal{G}(\mathcal{Y})$. 
\end{theorem}

While conditional independence tests give the theoretical result of a polynomial time recovery algorithm, in our experimental results we use the score-based approach, optimizing the BIC score which is efficient for trees using the Chow-Liu algorithm \citep{chow1968approximating} and for higher-order graphs using a cutting-plane algorithm \citep{gobnilp}.

%% file: tex/factor_loadings.tex
\subsection{Learning factor loadings, $P(\mathcal{X} | \mathcal{Y})$}
\label{sec:loadings}

In this section we describe a moments-based approach to recovering the factor loadings in the Anchored Discrete Factor Analysis model (step 3 of Algorithm~\ref{alg:overall}). We show how to make use of the recovered low-order moments and the structure learned in Section~\ref{sec:structure} to estimate the conditional distributions of the observations conditioned on the latent variables.

In this work, we use a noisy-or parametrization, a widely-used parametrization for Bayesian networks in medical diagnosis \citep[e.g.][]{Miller86,wood06} and binary factor analysis~\cite{Singliar}. In particular, we assume that the distribution of any observation conditioned on the latent variables has the form:
$P(x_j=0|y_1, ..., y_m) = (1-l_{j}) \prod_{i} f_{i,j}^{y_i}$. The parameters $f_{i,j}$ are known as {\em failure probabilities} and the $l_{j}$ parameters are known as {\em leak} probabilities.  Intuitively, the generative model can be described as follows: For every observation, $X_j$, each parent, $Y_i$, that is ``on'' has a chance to turn it on and fails with probability $f_{i,j}$. The leak probability $l_j$ represents the probability of $X_j$ being turned on without any of its parents succeeding.

%This parametrization is not necessary, but it is an interesting example of a binary conditional distribution that can be easily estimated from its low order moments. Linear Gaussian link functions for continuous variables (i.e. $P(X_j|y) = \mathcal{N}(\sum_i \alpha_{i,j}y_i, \sigma^2)$) would also serve in this context.
We use $\mathcal{G}(\mathcal{Y})$ to denote the Bayesian network structure over the latent variables that describes $P(\mathcal{Y})$. This graph structure is learned in section~\ref{sec:structure}.

To properly estimate a failure probability $f_{i,j}$ from low order moments, we need to separate the direct effect of $Y_i$ in turning on $X_j$ and all other indirect effects (e.g. Conditioning on $Y_i=1$ changes the likelihood of another latent variable $Y_{i^\prime}$ being on, which in turn affects $X_j$).

{\bf Estimation using Markov blanket conditioning:} The first method of separating direct and indirect effects uses the entire Markov blanket of each latent variable. Denote the Markov blanket of $Y_i$ in $\mathcal{G}(\mathcal{Y})$ as $B_i \subset \mathcal{Y}$. For any setting of $B_i = b$, the following is a consistent estimator of $f_{i,j}$:
\begin{equation}
\label{eq:markov}
\hat{f}_{i,j}^{blanket} = \frac{\hat{P}(x_j^{(0)}|y_i^{(1)}, b)}{\hat{P}(x_j^{(0)}|y_i^{(0)}, b)},
\end{equation}

as shown in the supplementary materials. 
The simplest application of this estimator is in models where the latent variables are assumed to be independent \citep[e.g.][]{Miller86,JerHalSon_nips13}, where the Markov blanket is empty and the estimator is simply
\begin{equation}
\label{eq:fdirect}
\hat{f}_{i,j}^{direct} = \frac{\hat{P}(x_j^{(0)}|y_i^{(1)})}{\hat{P}(x_j^{(0)}|y_i^{(0)})}.
\end{equation}
Unfortunately, for general graphs, in order to estimate these moments, we have to form moments that are of the same order as the Markov blanket of each latent variable which can potentially be quite large, even in simple tree models, giving poor computational and statistical efficiency. 

\begin{figure}[tb]
    \centering
    \includegraphics[scale=0.4]{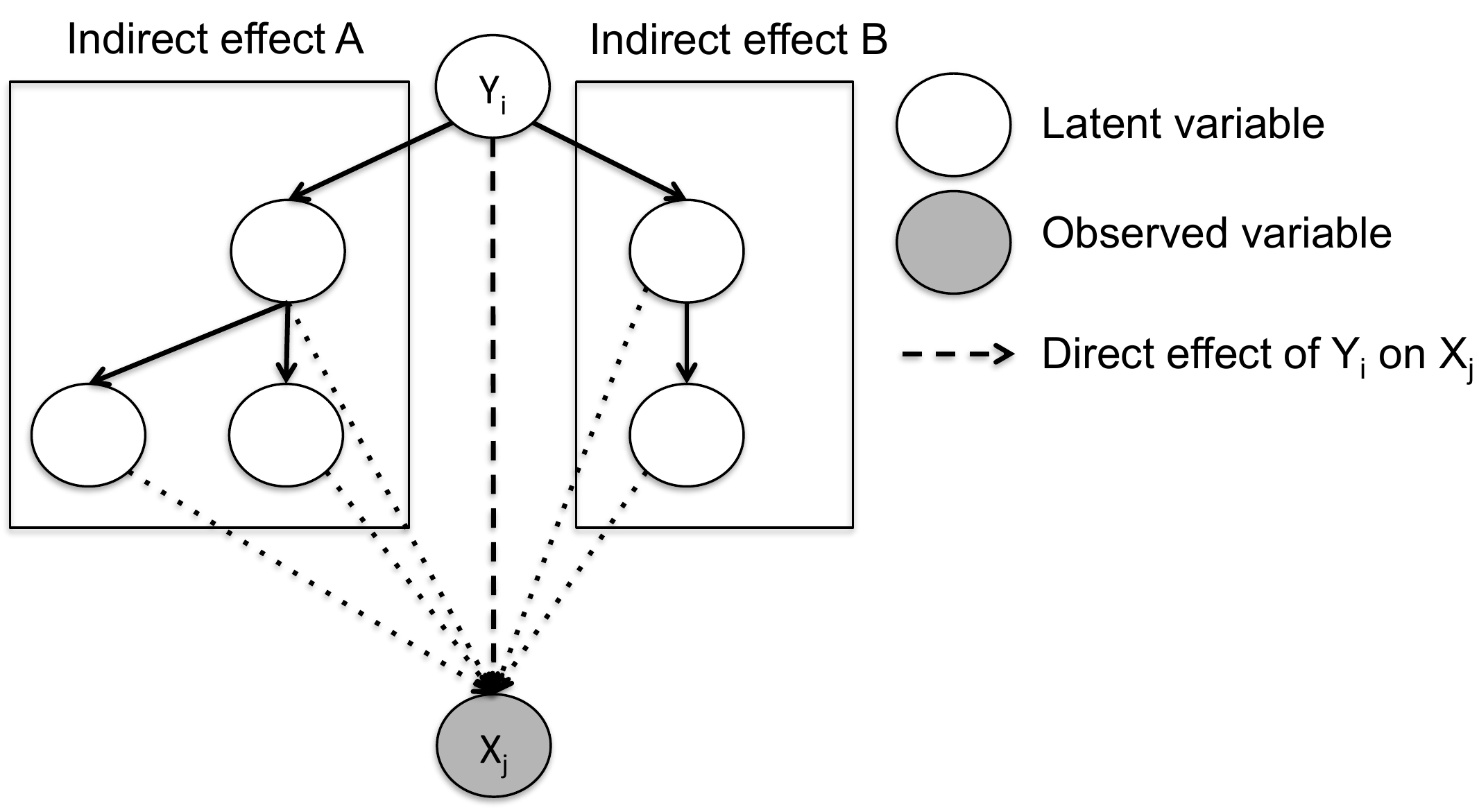}
    \caption{\label{fig:indirect}$Y_i$ has a direct effect on the distribution of $X_j$ as well as indirect effects that pass through its two neighbors. Correction factors are introduced to cancel the indirect effects through each neighbor, leaving only the direct effect which is modeled with the parameter $f_{i,j}$. }
\end{figure}

{\bf Improved estimation for tree-structured models:}
When the $\mathcal{G}(\mathcal{Y})$ is a tree, it is possible to correct for indirect effects serially, only requiring conditioning on two latent variables at a time. 
Each of the correction factors essentially cancels out the effect of an entire subtree in determining the likelihood that $X_j=1$.
Let $\mathcal{N}_i$ denote the neighbors of $Y_i$ in $\mathcal{G}(\mathcal{Y})$. 
When calculating $f_{i,j}^{tree}$, we introduce correction factors, $c_{i,j,k}$, to correct for indirect effects that flow through $Y_k \in \mathcal{N}_i$ (Figure~\ref{fig:indirect}). 

By correcting for each neighbor in serial we get the following estimator:
\begin{equation}
\label{eq:ftree}
\hat{f}_{i,j}^{tree} = \left(\prod_{k \in \mathcal{N}_i} \frac{1}{c_{i,j,k}}\right) \frac{\hat{P}(x_j^{(0)}|y_i^{(1)})}{\hat{P}(x_j^{(0)}|y_i^{(0)})},  
\end{equation}

where $c_{i,j,k}$ has the form: 
\begin{equation}
\label{eq:main_cdef}
c_{i,j,k} = \frac{\sum_{y_k} P(y_{k} | y_{i}^{(1)}) P(x_{j}^{(0)}|y_i^{(0)}, y_k)}
{P(x_{j}^{(0)} | y_{i}^{(0)})},
\end{equation} 
and can be estimated from the low-order moments recovered in Equation~\ref{eq:anchor_opt}. 
We note that for this method we only need to make assumptions about the graph of the latent variables, requiring no assumptions on the in-degree of the observations. For the correction factors to be defined, we do require that it is possible to condition on $(y_i=0, y_k=\{0,1\})$, so we require that $P(Y_k|Y_i)$ is not deterministic for all pairs of latent variables, $(Y_i,Y_k)$. 
A full derivation is provided in the supplementary materials.

\subsection{Estimating leak parameters}
Leak parameters can be estimated by comparing the empirical estimate of $\hat{P}(x_j^{(0)})$ to the one predicted by the model without leak, denoted as $P_{-l_j}(x_j^{(0)})$.

\begin{equation}
l_j = 1- \frac{P_{-l_j} (x_j^{(0)})}{\hat{P}(x_j^{(0)})}.
\end{equation}

If the latent variables are independent, the Quickscore algorithm~\citep{Heckerman_Quickscore} gives an efficient method of computing marginal probabilities:
\begin{equation}
\label{eq:quickscore}
P_{-l_j}(x_j^{(0)}) = \prod_{i} \left(P(y_i^{(0)}) + P(y_i^{(1)})f_{i,j}\right).
\end{equation}

For trees, this value can be computed efficiently using belief propagation and in more complex models it can be estimated by forward sampling.

\subsection{Putting it all together}

The full Anchored Factor Analysis model is a product of the latent distribution $P(\mathcal{Y})$, which is described by an arbitrary Bayesian network $G(\mathcal{Y}; \theta)$ and the factor loadings which are described by noisy-or link functions. The computational complexity of learning the model depends on the choice of constraints for the Bayesian network, $G(\mathcal{Y})$.
Table~\ref{table:complexity} outlines the different classes of models that can be learned and the associated computational complexities. We note that for some Bayesian network families, the complexity may be hard in a worst-case analysis, but practically cutting plane approaches that use integer linear programs can be successful in learning these models exactly and quickly \citep{gobnilp}. 

\begin{table*}
\centering
\small
\begin{tabular}{ l l l  c c}
\rowcolor{lightgray}
$\mathcal{G}(\mathcal{Y})$ & Complexity of learning $\mathcal{G}(\mathcal{Y})$ & Factor loadings \\ \hline
independent & None & $\hat{f}^{direct}$ (Eq. \ref{eq:fdirect}) &   \\
tree & Chow-Liu: $O(n^2)$ & $\hat{f}^{tree}$ (Eq. \ref{eq:ftree})  \\
degree-K & Independence tests: $O(n^2 \sum_{i=1}^K \binom{n}{i}2^{K})$ & $\hat{f}^{blanket}$ (Eq. \ref{eq:markov})  \\
indegree-K & Score-based: NP-hard worst case & $\hat{f}^{blanket}$ (Eq. \ref{eq:markov})   \\
\end{tabular}
\caption{\label{table:complexity} Complexity of learning different model classes. After performing the moment-transformations (step 1 in Algorithm~\ref{alg:overall}), the complexity of learning the models with latent variables is no harder than learning with fully observed moments.}
\end{table*}

%% file: tex/experiments.tex
%auto-ignore
%!TEX root = ../main.tex
\section{Experiments}
\label{sec:experiments}

We perform an empirical study of our learning algorithm using two real-world multi-tag datasets, testing its ability to recover the structure and parameters of the distribution of tags and observed variables.
\subsection{Datasets}

{\bf Medical records:} 
We use a corpus of medical records, collected over 5 years in the emergency department of a Level I trauma center and tertiary teaching hospital in a major city.
A physician collaborator manually specified 23 medical conditions that present in the emergency department and provided textual features that can be used as anchors as well as administrative billing codes that can be used for our purposes as ground truth. Features consist of binary indicators from processed medical text and medication records. 
Details of the processing and a selection of the physician-specified anchors are provided in the supplementary materials. 
Patients are filtered to exclude patients with fewer than two of the specified conditions, leaving 16,268 patients.

{\bf Stack Overflow:} We also evaluate our methods on a large publicly available dataset using questions from Stack Overflow\footnote{\url{http://blog.stackoverflow.com/category/cc-wiki-dump/}. This data used was also used in a Kaggle competition: \url{https://www.kaggle.com/c/facebook-recruiting-iii-keyword-extraction}}, we treat user provided tags as latent variables. 
The observed vocabulary consists of the 1000 most common tokens in the questions, using a different vocabulary for the question header and body. 
Each question is described by a binary bag-of-words feature vector, denoting whether or not each word in the vocabulary occurs in the question text. 
We use a simple rule to define anchor variables: for each tag, the text of the tag appearing in the header of the question is used as an anchor. 
For example, for the tag {\em unix}, the anchor is a binary feature indicating whether the header of the text contains the word {``unix''}. We use the 50 most popular tags in our models.

\subsection{Methods}

In all of our experiments, we provide the algorithms and baselines with the empirical conditional probabilities of the anchors. Other methods of estimating these values exist and are discussed in Section~\ref{sec:noise_rates}, but here we use the ground truth values for these noise rates in order to focus on the errors that arise from modeling error and finite data samples. 
Details on the optimization procedures and parameters used can be found in the supplementary materials. 
%and code is available to recreate our results on the public dataset.

\subsection{Latent variable representation}

\begin{figure*}[t]
\centering
    \includegraphics[scale=0.5]{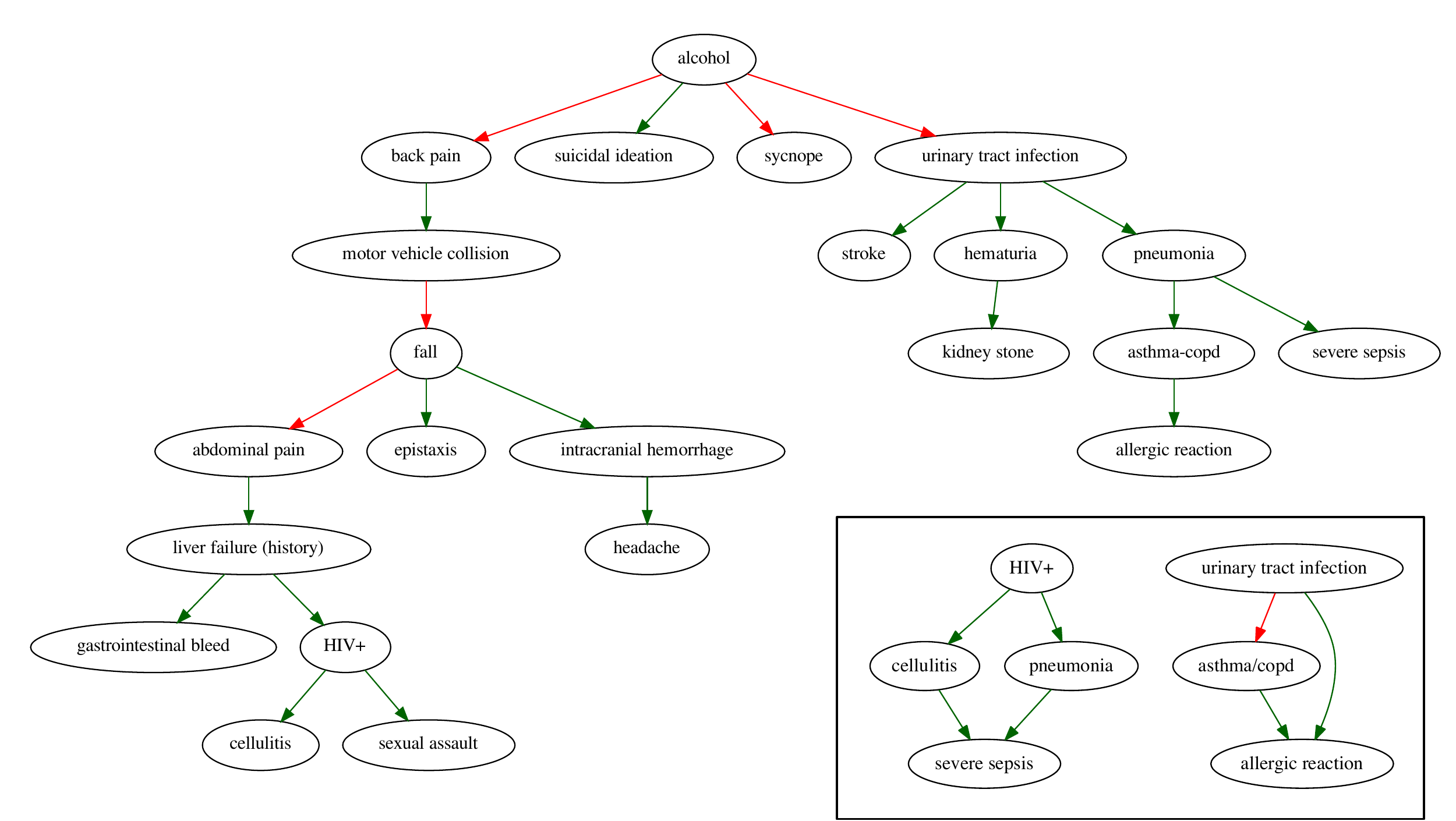}
    \caption{\label{fig:tree}\small
The learned tree-structured latent variable model describing the distribution of 23 conditions in the emergency department, learned using marginal polytope constraints. 
Green lines represent positive correlations and red lines represent negative correlations. 
The section in the box shows small subgraphs of a more complex structure learned allowing for two parents per variable.
}
\end{figure*}

In this section we evaluate the quality of the learned representations of the latent variables, $P(\mathcal{Y})$. This task in interesting in its own right in settings where we care about understanding the interactions of the latent variables themselves for the purposes of knowledge discovery and understanding.
Figure~\ref{fig:tree} shows a tree-structured graphical model learned to represent the distribution of the 23 latent variables in the Emergency dataset as well as small sub-graphs from a more complex model. 

In the tree-structured model, highly correlated conditions are linked by an edge. For example, asthma and allergic reactions, or alcohol and suicidal ideation. This is significant, since the model learning procedure learns only using anchors and without access to the underlying conditions.

The insert shows subgraphs from a model learned with two parents per variable. 
This allows for more complex structures. For example: being HIV positive makes the patient more at risk for developing infections, such as cellulitis and pneumonia. Either one of these is capable of causing septic shock in a patient.
The v-structure between cellulitis, septic shock, and pneumonia expresses the {\em explaining away} relationship: knowing that a patient has septic shock, raises the likelihood of cellulitis or pneumonia. 
Once one of those possible parents is discovered (e.g., the patient is known to have cellulitis), the septic shock is explained away and the the probability of having pneumonia is reduced. 
In the second example relationship, both asthma and urinary tract infections are associated with allergic reactions (asthma and allergic reactions are closely related and many allergic reactions in hospital occur in response to antibiotics administered for infections), but asthma and urinary tract infections are negatively correlated with each other since either one is sufficient to explain away the patient's visit to the emergency department.

\subsubsection{Robustness to model error}
In practice, the anchors that we specify are never perfect anchors, i.e., they don't fully satisfy the conditional independence conditions of Definition~\ref{def:anchor}.
We find that enforcing marginal or local polytope constraints (see section~\ref{sec:robust}) during the moment recovery process provides moment estimates that are more robust to imperfect anchors, improving the overall quality of the learned models. 

\begin{figure}
\centering
    \includegraphics[scale=0.27]{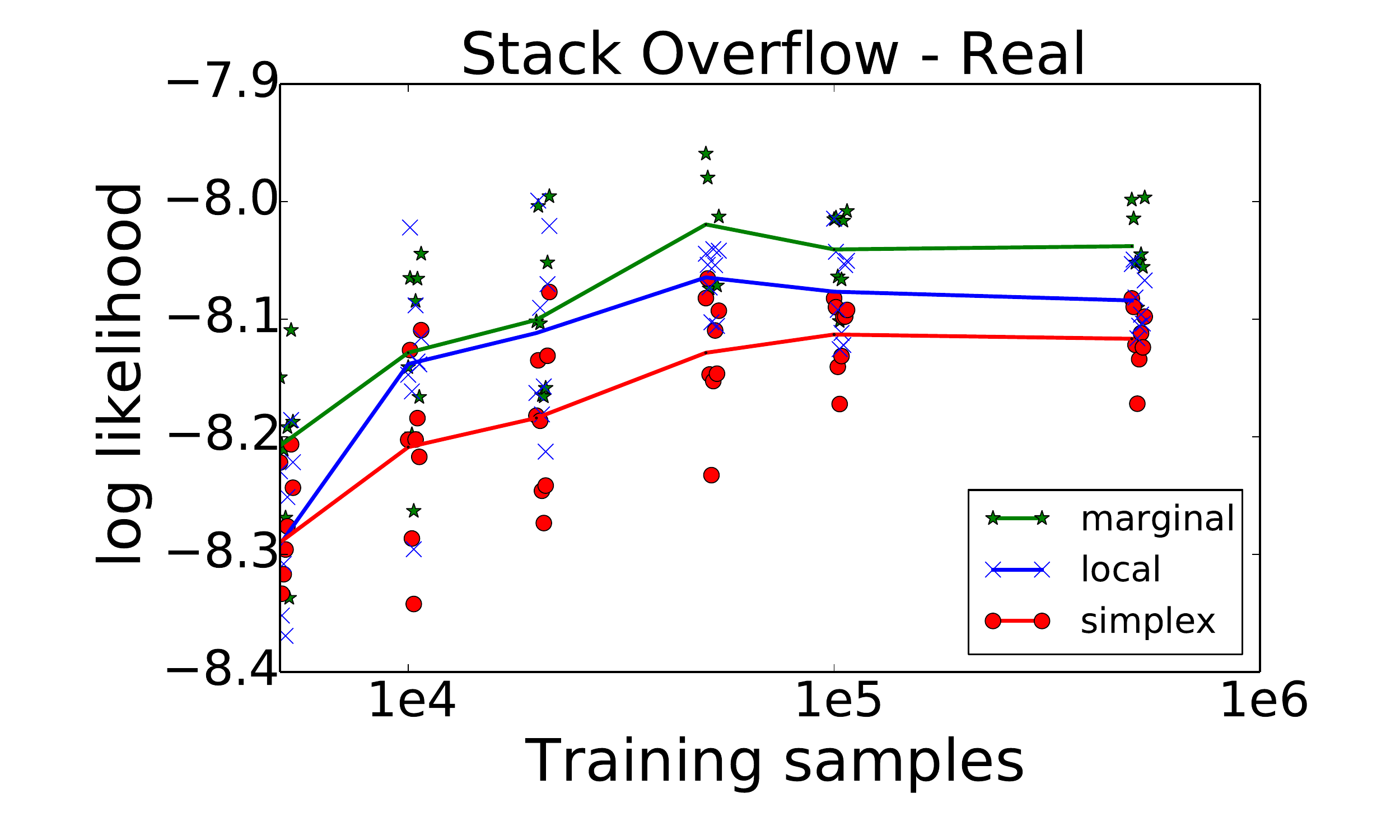}
    \includegraphics[scale=0.27]{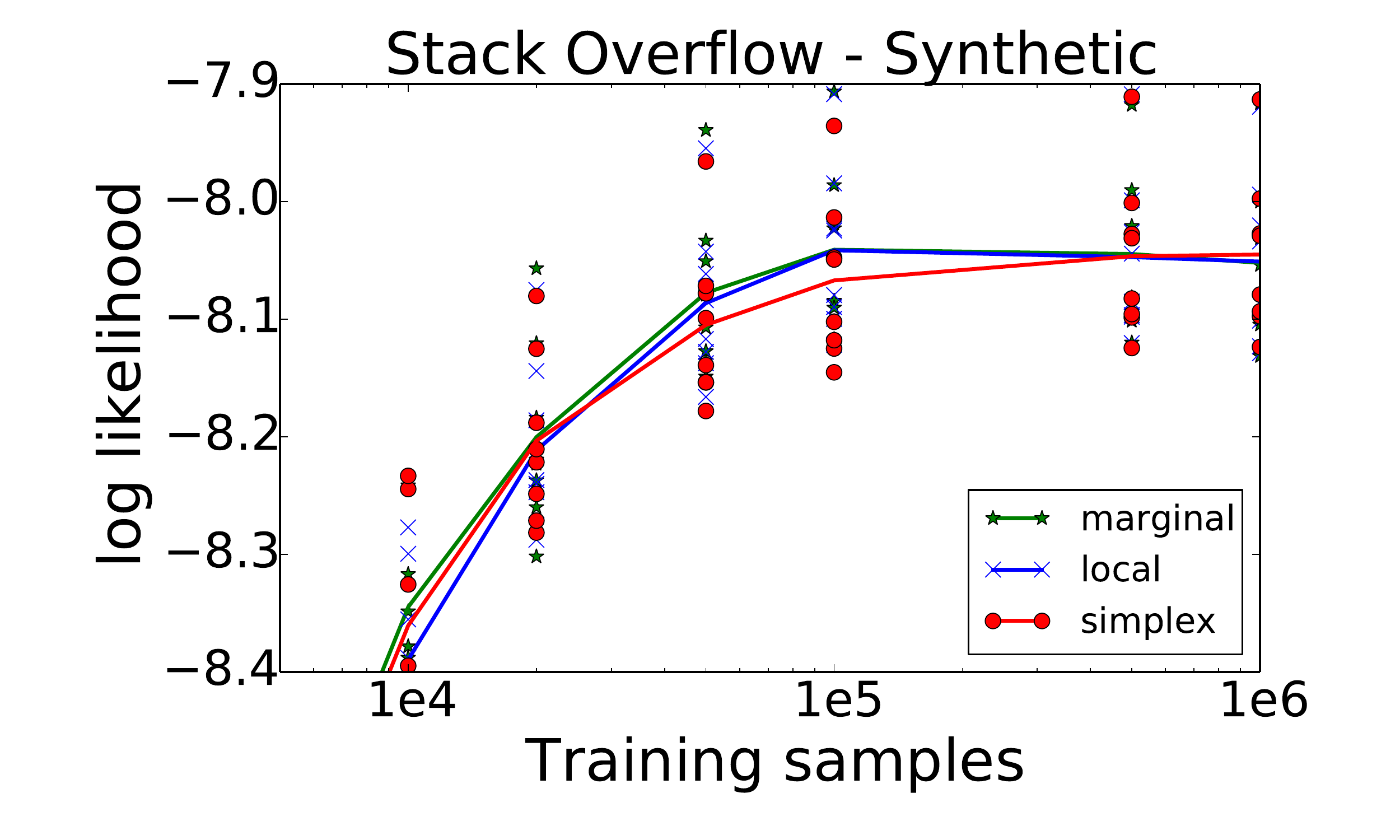}
    
\caption{\label{fig:heldout-factors}\small
(Left) Per document held-out likelihood of learned tree structured models on Stack Overflow (50 variables) using held-out sets of 10K documents, each marker represents a different random train/test sample for the data. Solid lines represent the average of 8 runs.  Different lines represent successively tight outer bounds to the marginal polytope constraints. (Right) When the Stack Overflow data is replaced by synthetic data drawn from a learned model, the difference between the constraints is much less pronounced.}
\end{figure}

The weakest constraint we consider is using independent simplex constraints as in \cite{chaganty2014graphical}. In addition, we evaluate models learned with moments recovered using local consistency constraints, described in Equation~\ref{eq:local_consistency} and the marginal polytope constraints described in Section~\ref{sec:robust}. 
Figure~\ref{fig:heldout-factors} shows the held-out likelihood of the tags according to tree models learned from moments recovered with increasingly tight approximations to the marginal polytope constraints on the Stack Overflow corpora. The tighter constraints learn higher quality models. This effect persists even in the large sample regime, suggesting that the residual gaps between the methods are due to sensitivity to model error. On synthetically generated corpora, this effect disappears and all three constrained optimization approaches perform equally well.

Solving with tighter constraints does increase running time. 
For example, using our current solving procedures, learning a model with 50 latent variables requires 30 seconds using the simplex constraints, 519 seconds for the local polytope and 6,073 seconds for the marginal polytope.
Unlike EM-based procedures for learning models with latent variables which iterate over training samples in an inner loop, the running time of these method-of-moments approaches depends on the number of samples only when reading the data for the first time.

\subsection{Factor analysis -- relating observations to latent variables}

Table~\ref{table:factors} shows a selection of learned factor loadings on the Stack Overflow dataset, learned using the tree-structured estimation procedure described in Section~\ref{sec:loadings}. The rest of the learned factors for both Stack Overflow and the Emergency datasets are found in the supplementary materials.

\begin{figure}[t]
\centering
\includegraphics[scale=0.3]{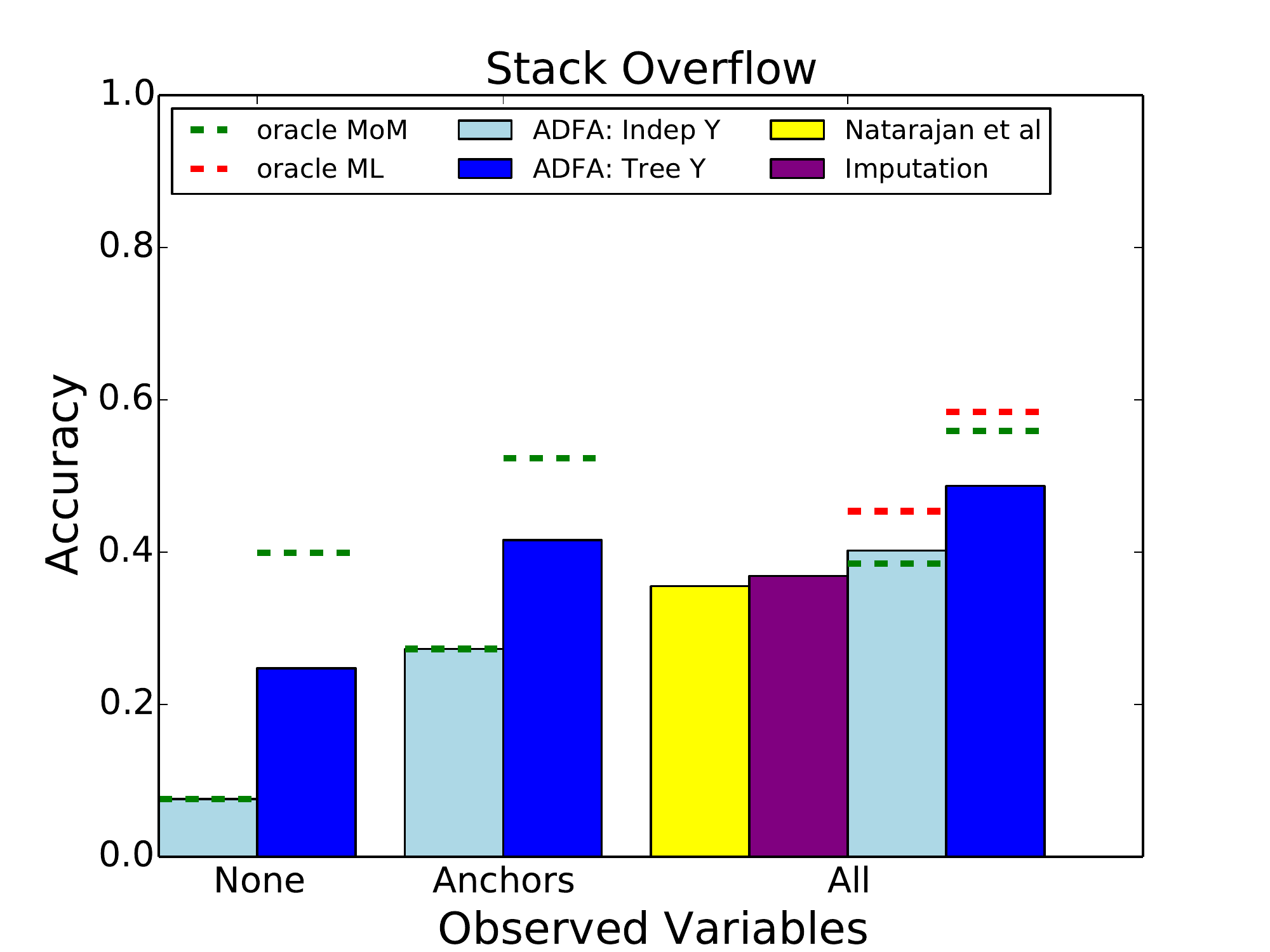}\hspace{-1ex}
\includegraphics[scale=0.3]{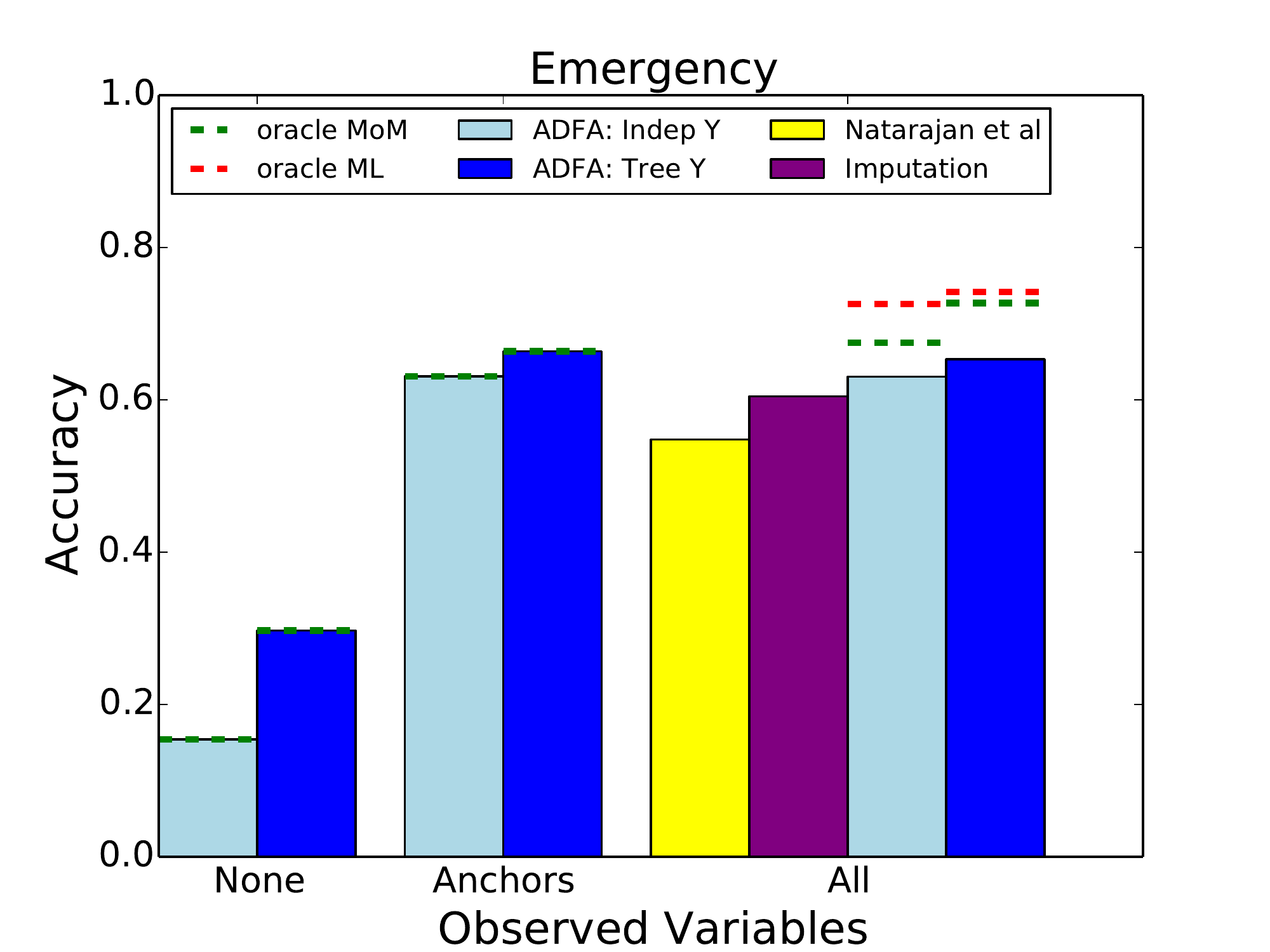}\hspace{-1ex}
\caption{\label{fig:lasttag}\small Average accuracy on the last-tag prediction task for 5K held-out instances for Emergency and Stack Overflow datasets. Anchored Discrete Factor Analysis (ADFA) models are learned using a tree structured model for the latent variables and an independent model.}
\vspace{-3mm}
\end{figure}

\subsection{Comparison to related methods}

We evaluate the full learned factor analysis model with a simulated last-tag prediction task where the model is presented with all but one of the positive tags (i.e. the latent variables) for a single document or patient and the goal is to predict the final tag. 
Our learned models use a tree-structured Bayesian network to represent the distribution $P(\mathcal{Y})$ and noisy-or gates to represent the conditional distributions $P(\mathcal{X}|\mathcal{Y})$.

{\bf Noise tolerant discriminative training:} \quad
As a baseline, we compare to the performance of a noise-tolerant learning procedure \citep{natarajan2013learning} using independent binary classifiers, trained using logistic regression with reweighted samples as described in their paper. To ensure a fair comparison with our learning algorithm which is provided with the anchor noise rates, we also provide the baselines with the exact values for the noise rates. 
Since the learning approach of \citep{natarajan2013learning} is not designed to use the noisy labels (anchors) at prediction time, we consider two variants: one ignores the anchors at test time, the other predicts only according to the noise rates of the anchors (ignoring all other observations), if the anchor is present. 
%In addition, since the anchors in the Emergency dataset can be considered noisy labels with very low false positive rates, we compare to the positive-unlabeled learning algorithm described in \citet{elkan2008learning}. 
%This method has previously been used for prediction in emergency department settings where ground truth labels are hard to obtain \citep{HalpernEtAl_amia14}.
The baseline that we compare against is the best of the baselines described above. 

{\bf Oracle comparisons:} \quad
We compare to two different oracle implementations to deconvolve different sources of error. 
The first, {\em Oracle MoM}, uses a method-of-moments approach to learning the joint model (as described in Sections~\ref{sec:structure} and \ref{sec:loadings}), but uses oracle access to the marginal distributions that include latent variables (i.e. does not incur any error when recovering these distributions using the method described in Section~\ref{sec:recovering}). 
The second oracle method {\em Oracle ML}, uses a maximum likelihood approach to learning the joint models with full access to the latent variables for every instance (i.e., learning as though all data is fully observed.) Comparing the gap between oracle ML and oracle MoM shows the loss in accuracy that comes from choosing a method-of-moments approach over a maximum likelihood approach. 
\begin{table}[t]
\tiny
\begin{tabular}{|c|l|}
\hline
\multicolumn{2}{|c|}{Stack Overflow}\\
\hline
{\bf Tag} & {\bf Top weights} \\
\hline
osx & osx, i'm, running, i've, install, installed, os, code \\
image & image, code, size, upload, html, save, picture, width \\
mysql & mysql, query, rows, row, id, 1, tables, join  \\
web-services & web, service, web-services, client, java, services \\
linux & linux, running, command, machine, system, server \\
query & query, table, a, result, results, queries, tables, return \\
regex & regex, match, expression, regular, pattern, i'm \\
\hline
\end{tabular}
\begin{tabular}{|c|l|}
\hline  
\multicolumn{2}{|c|}{Emergency}\\
\hline
{\bf Tag} & {\bf Top weights} \\
\hline
abdominal pain & pain, Ondansetron, nausea, neg:fevers \\
alcohol & sex:m, sober, admits, found, drink \\
asthma-copd & albuterol sulfate, sob, Methylprednisolone \\
stroke & age:80-90, admit, patient, head, ekg: \\
hematuria & sex:m, urine, urology, blood, foley \\
HIV+ & sex:m, Truvada, cd4, age:40-50, Ritonavir \\
motor vehicle collision & car, neg:loc, age:20-30, hit, neck \\
\hline
\end{tabular}
\caption{\label{table:factors} A selection of the learned factor loadings in the Stack Overflow and emergency datasets. Highly weighted words are words with low failure probabilities in the noisy-or parametrization.}
\end{table}

{\bf Imputation:} \quad
We compare to an imputation-based learning method, which learns a maximum likelihood model (Tree structured $P(\mathcal{Y})$ and noisy-or gates for conditional distributions $P(\mathcal{X}|\mathcal{Y})$ from the fully imputed data, using a single sample from the independent binary baseline classifiers \citep{natarajan2013learning} to impute the values of the latent variables. 

{\bf Models with independent latent variables:} \quad
Previous works \citep[e.g.,][]{JerHalSon_nips13} consider similar models where latent variables are assumed to be independent. 
In our work, we explicitly model dependencies between the latent variables. We test the advantage of having a structured representation of the latent variables compared to a model with independent latent variables. 

\begin{figure}[t]
\centering
\includegraphics[scale=0.33]{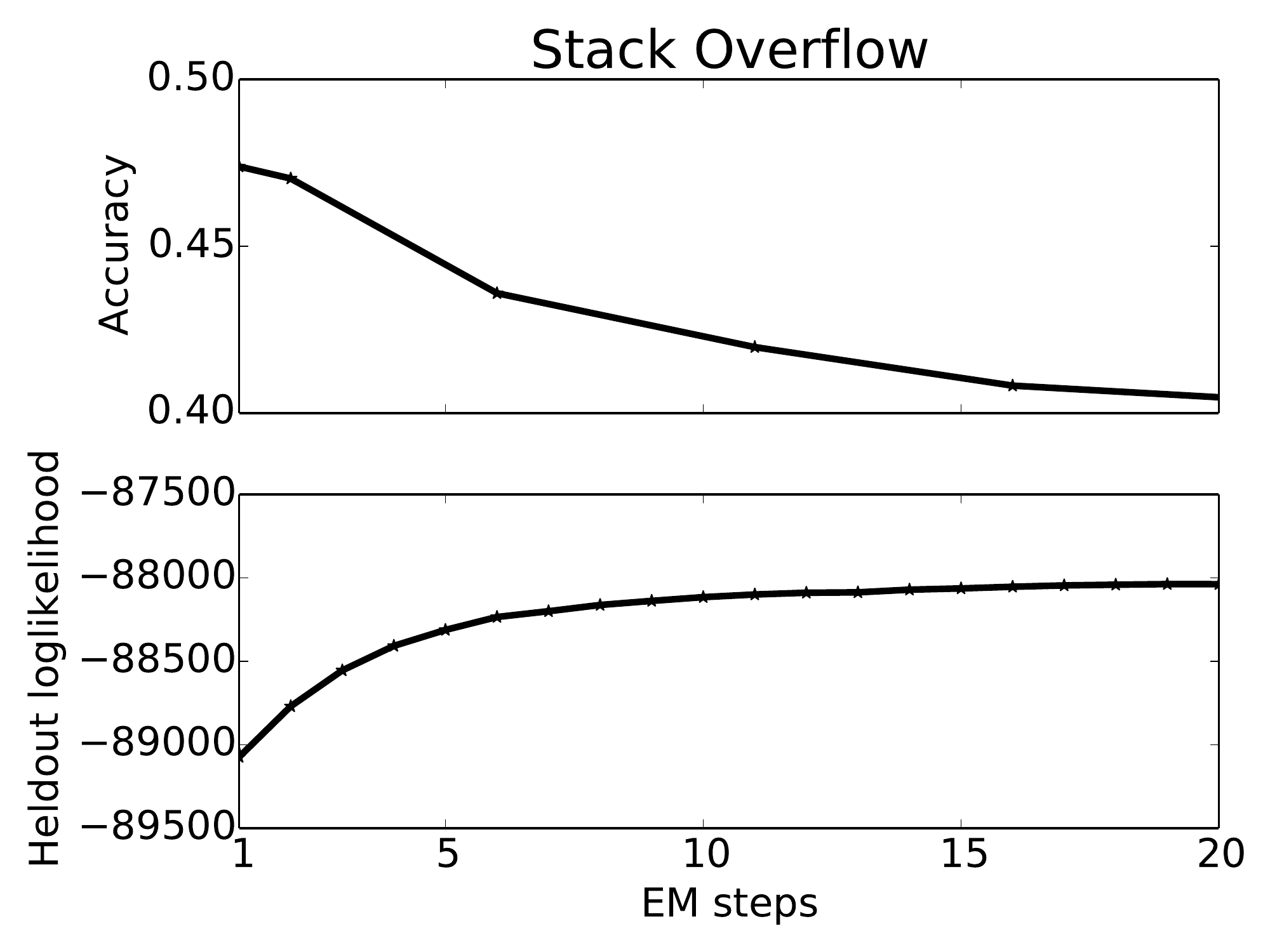}
\includegraphics[scale=0.33]{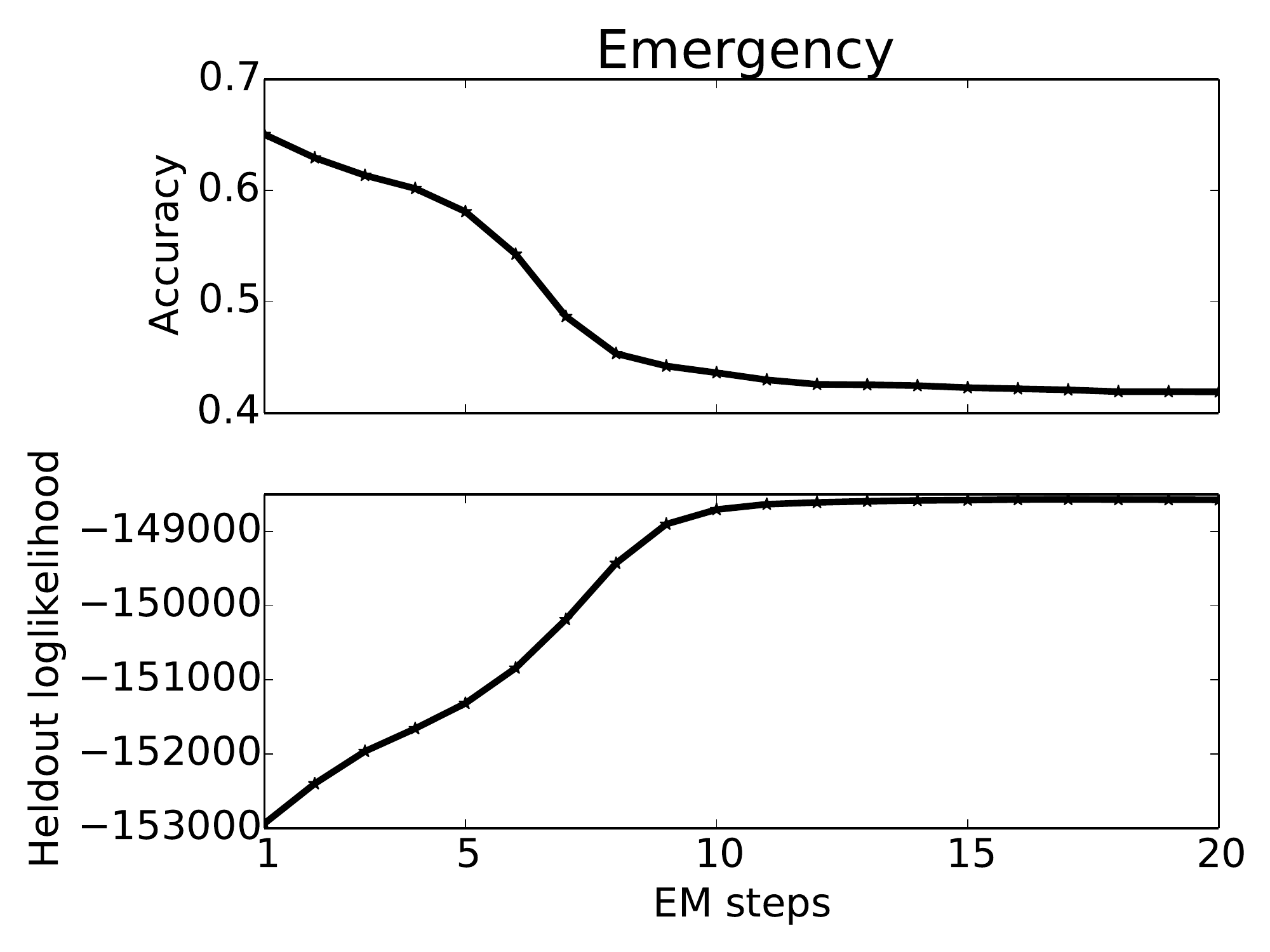}
\small
    \begin{tabular}{|l|l|}
    \hline
    \rowcolor{lightgray}
    \textbf{Before optimization} & \textbf{After 9 EM steps}  \\
    \hline
    anchor:headache & anchor:headache\\
    neg-changes & neg-chest \\
    nausea & neg-chest\_pain \\
    a & neg-ed \\
    neg-, & neg-breath \\
    neg-vision & neg-imaging\\
    neg-or & neg-shortness\\
    neg-weakness & neg-shortness\_breath\\
    head & neg-course\\
    neg-of & neg-interventions\\
    \hline
    \end{tabular}
\caption{\label{fig:degrading_em}\small (top) Effects of pure likelihood-based optimization on accuracy. 
A likelihood optimization procedure is initialized with the results of the anchor-based method-of-moments estimate for the Stack Overflow and Emergency datasets (ADFA-tree). 
As the likelihood optimization progresses, classification accuracy degrades.
(bottom) The meaning of the latent variable associated with ``headache'' changes over the course of likelihood optimization. Before optimization the highly weighted words are associated with the headache exam (e.g., head, changes of vision, nausea). After a small number of EM-steps, the meaning of the latent variable has changed and is now associated with long negation scopes.}
\end{figure}

{\bf Maximizing data likelihood:} \quad
We could learn a model without explicitly using the anchors, attempting to maximize likelihood of the observations. This maximization would be a non-convex problem due to the latent variables, and marginalizing the latent variables to compute the likelihood is computationally difficult. However, an EM procedure \citep{DLR} can be applied here (more detail in the supplementary materials). We initialize the parameters using the ADFA-tree method and run EM to determine whether additional steps of likelihood optimization can improve beyond the solutions found with ADFA.

\subsection{Results} Figure~\ref{fig:lasttag} shows the improvement over baseline on the heldout tag prediction task for the two real-world datasets.
We observe that learning the factor analysis model does indeed help with prediction in both tasks and we believe that this is because the independent classifiers in the baseline cannot take advantage of the correlation structure between latent variables. 
In the Stack Overflow corpus, we see that there is a clear advantage to learning a structured representation of the latent variables as the performance of the tree structured discrete factor analysis model outperforms the oracle bounds of a model with independent variables. The emergency corpus is difficult to improve upon because the anchors themselves are so informative.
In both datasets, the oracle results show that using method-of-moments for parameter learning is not a big source of disadvantage compared to maximum likelihood estimation.

Figure~\ref{fig:degrading_em} shows the effect of running likelihood-based optimization. Contrary to expectation, improving the likelihood of data {\em does not} improve performance on the tag-prediction task. One reason for this is that latent variables can change their meaning from the original intent as a result of the likelihood optimization. One example in Figure~\ref{fig:degrading_em} shows how the ``headache'' latent variable is used to model long negation scopes in patient records rather than modeling headaches.

Although the tighter constraints on the marginal polytope help for density estimation of $P(\mathcal{Y})$ and could be useful for knowledge discovery, for this particular task the improved density estimation does not result in significant changes in accuracy. Thus, we show only the results of the model learned with simplex constraints here.

%% file: tex/Discussion.tex
%auto-ignore
%!TEX root = ../main.tex
\section{Conclusion}
Learning interpretable models with latent variables is a difficult task. 
In this paper we present a fast and expressive method that allows us to learn models with complex interactions between the latent variables. 
The learned models are also interpretable due to the effect of the user-specified anchor observations. 
On real-world datasets, we show that modeling the correlations between latent variables is useful, and outperform competitive baseline procedures. We find that enforcing marginal polytope constraints is useful for improving robustness to model error, a technique we believe can be more widely applied. 
Anchors have an interesting property that they make the structure and parameter estimation  with latent variables as easy as learning with fully observed data for method-of-moments algorithms that require low order moments. 
In contrast, likelihood-based learning remains equally hard, to the best of our knowledge, even with anchors.

%% file: tex/Acknowledgments.tex
%auto-ignore
%!TEX root = ../main.tex
\section{Acknowledgments}
This material is based upon work supported by the National Science Foundation under Grant No. 1350965 (CAREER award). Research is also partially support by an NSERC postgraduate scholarship.

%% file: tex/supplemental_materials.tex
%auto-ignore
%!TEX root = ../main.tex

\section{R matrix is full rank}
\label{append:full_rank}
In this section we show that the matrix $R$, defined in Section~\ref{sec:exclusiveviews} is invertible. 
$R$ is block-diagonal, so its determinant is equal to the product of the determinants of the blocks and will be non-zero as long as all of the blocks have non-zero determinants.

Each block, $R_{\mathcal{Z}}$ is a Kronecker product of conditional distributions: $R_{\mathcal{Z}} = \otimes_{k=1}^{|\mathcal{Z}|} P(A_k | Z_k)$.
The determinant of each of these $2\times 2$ matrices is nonzero as long as $P(A_k | Y_k=0) \neq P(A_k | Y_k=1)$ which is assumed in the definition of anchored latent variable models since $A_{Y_k} \not \perp Y_k$.Thus, the determinant of the Kronecker product is also non-zero, and the $R$ matrix is full rank.

\section{Estimating conditional probabilities with two anchors per latent variable}
In section~\ref{sec:noise_rates}, we note that if a latent variable as two anchors, their conditional distributions can estimated. The derivation is as follows:
Let $W_1, W_2$ be observed anchors of $Y_i$.
We can choose any other observation $X_j$ that has positive mutual information with both $W_1$ and $W_2$. In that case, we have a {\em singly-coupled triplet} setting, as described in \cite{HalpernSontag_uai13}. 
Marginalizing over all latent variables but $Y_i$, we have that the observed distribution $P(W_1, W_2, X_j)$ is a $2\times2\times2$ tensor with the following decomposition into a sum of rank-1 tensors:

\begin{align}
P(W_1, W_2, X_j)  = &P(Y_i=0)P(W_1|Y_i=0)P(W_2|Y_i=0)P(X_j|Y_i=0)  \notag \\
&+ P(Y_i=1)P(W_1|Y_i=1)P(W_2|Y_i=1)P(X_j|Y_i=1)
\end{align}
This decomposition can be computed efficiently \cite{Berge}, and the conditional probabilities can be recovered.

\section{Estimating failures with Markov blanket conditioning}
The Markov Blanket estimator from Section~\ref{sec:loadings} follows from the noisy-or parametrization of the model. Here we show that the estimator from Equation~\ref{eq:markov} is indeed a consistent estimator. We start with the estimator:

\begin{equation*}
f_{i,j}^{blanket} = \frac{P(X_j=0|Y_i=1, B=b)}{P(X=0|Y_i=0, B=b)},
\end{equation*}

Breaking the numerator and denominator of the RHS. Let $\mathcal{C} = B \cup Y_i$ represent the conditioned variables and $\mathcal{U} = \mathcal{Y} \setminus \mathcal{C}$ be the unconditioned variables. For the numerator, we have:

\begin{align}
P(X_j=0|Y_i=1, B=b) &= (1-l_j) f_{i,j} \prod_{i^\prime \in B} f_{i^\prime, j}^{y_{i^\prime}} \left(\sum_{y_{\mathcal{U}}}P(y_{\mathcal{U}}|Y_i=1, B=b)\prod_{k \in \mathcal{U}} f_{k, j}^{y_{k}}\right) \notag \\
& = f_{i,j} \prod_{i^\prime \in B} f_{i^\prime, j}^{y_{i^\prime}} \left(\sum_{y_{\mathcal{U}}}P(y_{\mathcal{U}}|B=b)\prod_{k \in \mathcal{U}} f_{k, j}^{y_{k}}\right).
\end{align}

Similarly, for the denominator:
\begin{align}
P(X_j=0|Y_i=0, B=b) &= (1-l_j) \prod_{i^\prime \in B} f_{i^\prime, j}^{y_{i^\prime}} \left(\sum_{y_{\mathcal{U}}}P(y_{\mathcal{U}}|Y_i=0, B=b)\prod_{k \in \mathcal{U}} f_{k, j}^{y_{k}}\right) \notag \\
& = \prod_{i^\prime \in B} f_{i^\prime, j}^{y_{i^\prime}} \left(\sum_{y_{\mathcal{U}}}P(y_{\mathcal{U}}|B=b)\prod_{k \in \mathcal{U}} f_{k, j}^{y_{k}}\right).
\end{align}

The second lines follow from the Markov blanket property. Thus, the ratio is equal to $f_{i,j}$. As the number of samples approaches infinity, the empirical estimates, $\hat{P}(X_j=0|Y_i=0, B=b)$ and $\hat{P}(X_j=0|Y_i=1, B=b)$, respectively approach the values of the true probabilities, and thus the estimator is consistent, provided that, $P(X_j=0|Y_i=0, B=b) > 0$.

\section{Estimating correction coefficients serially in trees}
In this section we derive the correction terms for estimating the failure probabilities $f_{i,j}$ in tree models described in Section~\ref{sec:loadings} (Equation~\ref{eq:ftree}). While the directionality of the tree is not important, it is easier notationally to consider a rooted tree, and without loss of generality, when estimating the failure term $f_{i,j}$, we can consider the tree as rooted at $Y_i$. 

We introduce the notation $P_{\mathcal{T}(Y_i)}(e)$ to denote the likelihood of an event $e$ in the graphical model where all nodes but $Y_i$ and its non-descendants are removed (i.e., all that remains is the subtree rooted at $Y_i$).

Conditioning on the variable $Y_i$ taking the value $y_i$, the likelihood of $X_j=0$ is written as:
\begin{align}
\label{eq:single_cond}
P(X_j = 0 | y_i) &= (1-l_j)P_{\mathcal{T}(Y_i)}(X_j=0 | y_i) \notag \\
                 &= (1-l_j)f_{i,j}^{y_i} \prod_{k \in \text{child}(Y_i)} \sum_{y_k} P(y_k | y_i) P_{\mathcal{T}(Y_k)}(X_j=0 | y_k).
\end{align}

Correction terms are defined as 
\begin{equation}
c_{i,j,k} = \frac{\sum_{y_k} P(y_k | y_i^{(1)}) P_{\mathcal{T}(Y_k)}(X_j=0 | y_k)}
{\sum_{y_k} P(y_k | y_i^{(0)}) P_{\mathcal{T}(Y_k)}(X_j=0 | y_k)}.
\end{equation}

Using Equation~\ref{eq:single_cond}, it is easy to see that:
\begin{equation}
\frac{P(X_j = 0 | y_i^{(1)})}{P(X_j = 0 | y_i^{(0)})} \prod_{k \in \text{child}(Y_i)} \frac{1}{c_{i,j,k}} = f_{i,j}.
\end{equation}

It remains to be shown that $c_{i,j,k}$ can be estimated from low order moments as in Equation~\ref{eq:main_cdef}.
\begin{equation}
\label{eq:cdef}
c_{i,j,k} = \frac{\sum_{y_k} P(y_{k} | y_{i}^{(1)}) P(x_{j}^{(0)}|y_i^{(0)}, y_k)}
{P(x_{j}^{(0)} | y_{i}^{(0)})}.
\end{equation}

We can expand $P(x_{j}^{(0)}|y_i^{(0)}, y_k)$ as follows:

\begin{align}
\label{eq:double_condition}
P(&x_{j}^{(0)}|y_i^{(0)}, y_k) \notag \\
&= (1-l_{j})f_{k,j}^{y_k} \prod_{k^\prime \in \text{child}(Y_i) \setminus k} \sum_{y_{k^\prime}} P(y_{k^\prime} | y_i^{(0)}, y_k) P_{\mathcal{T}(Y_{k^\prime})}(X_j=0 | y_{k^\prime}) \notag \\
& \times \prod_{k^\prime \in \text{child}(Y_k)} \sum_{y_{k^\prime}} P(y_{k^\prime} | y_i^{(0)}, y_k)
P_{\mathcal{T}(Y_{k^\prime})}(X_j=0 | y_{k^\prime}) \notag \\
& = (1-l_{j})f_{k,j}^{y_k}\prod_{k^\prime \in \text{child}(Y_i) \setminus k} \sum_{y_{k^\prime}} P(y_{k^\prime} | y_i^{(0)}) P_{\mathcal{T}(Y_{k^\prime})}(X_j=0 | y_{k^\prime}) \notag \\
& \times \prod_{k^\prime \in \text{child}(Y_k)} \sum_{y_{k^\prime}} P(y_{k^\prime} | y_k)
P_{\mathcal{T}(Y_{k^\prime})}(X_j=0 | y_{k^\prime})
\end{align}

Where the second equality comes from the conditional independence properties that conditioning on $(Y_i,Y_k)$ makes children of $Y_k$ independent of $Y_i$ and children of $Y_i$ (other than $Y_k$) independent of $Y_k$.

\vspace{2em}
We now substitute Eq.~\ref{eq:double_condition} into Eq.~\ref{eq:cdef} and treat the numerator (num) and denominator (denom) separately:
\begin{align}
\text{num} &= (1-l_{j})\sum_{y_k} P(y_{k} | y_{i}^{(1)}) f_{k,j}^{y_k} \prod_{k^\prime \in \text{child}(Y_k)} \sum_{y_{k^\prime}} P(y_{k^\prime} | y_k) P_{\mathcal{T}(Y_{k^\prime})}(X_j=0 | y_{k^\prime}) \notag \\
& \times \prod_{k^\prime \in \text{child}(Y_i) \setminus k} \sum_{y_{k^\prime}} P(y_{k^\prime} | y_i^{(0)}) P_{\mathcal{T}(Y_{i^\prime})}(X_j=0 | y_i^{(0)})
\end{align}

The product over children of $Y_i$ does not depend on $y_k$ and can be pulled out of the sum.

\begin{align}
\text{num}  &= (1-l_{j})\prod_{k^\prime \in \text{child}(Y_i) \setminus k} \sum_{y_{k^\prime}} P(y_{k^\prime} | y_i^{(0)}) P_{\mathcal{T}(Y_{i^\prime})}(X_j=0 | y_i^{(0)})\notag \\
& \times \sum_{y_k} P(y_{k} | y_{i}^{(1)}) f_{k,j}^{y_k}
\prod_{k^\prime \in \text{child}(Y_k)} \sum_{y_{k^\prime}} P(y_{k^\prime} | y_k) P_{\mathcal{T}(Y_{k^\prime})}(X_j=0 | y_{k^\prime}) \notag \\
& =(1-l_{j})\prod_{k^\prime \in \text{child}(Y_i) \setminus k} \sum_{y_{k^\prime}} P(y_{k^\prime} | y_i^{(0)}) P_{\mathcal{T}(Y_{i^\prime})}(X_j=0 | y_i^{(0)}) \notag \\
&\times \sum_{y_k} P(y_{k} | y_{i}^{(1)}) P_{\mathcal{T}(Y_k)}(X_j=0 | y_k)
\end{align}

Expanding the denominator using Equation~\ref{eq:single_cond}:
\begin{equation}
\text{denom} = P(x_{j}^{(0)} | y_{i}^{(0)}) = (1-l_j)\prod_{k^\prime \in \text{child}(Y_i)} \sum_{y_k^\prime} P(y_{k^\prime} | y_i^{(0)}) P_{\mathcal{T}(Y_{k^\prime})}(X_j=0 | y_{k^\prime})
\end{equation}

Canceling the leak term and the product over all children of $Y_i$ except for $Y_k$ from both the numerator and denominator, we are left with:

$$
\frac{\text{num}}{\text{denom}} = \frac{\sum_{y_k} P(y_{k} | y_{i}^{(1)}) P_{\mathcal{T}(Y_k)}(X_j=0 | y_k)}{\sum_{y_k} P(y_k | y_i^{(0)}) P_{\mathcal{T}(Y_k)}(X_j=0 | y_k)}
$$
as desired.
% $$
% \frac{P(y_{k}^{(0)} | y_{i}^{(1)}) P(x_{j}^{(0)}|y_i^{(0)}, y_k^{(0)}) + P(y_{k}^{(1)} | y_{i}^{(1)}) P(x_{j}^{(0)}|y_i^{(0)}, y_k^{(1)})}
% {P(x_{j}^{(0)} | y_{i}^{(0)})}
% $$

\section{Dataset preparation}
In this section we provide additional details about the preparation of the two real-world datasets used in the experimental results.

{\bf Emergency:} The emergency dataset consists of the following fields from patients' medical records: Current medications (medication reconciliation) and Administered medications (pyxis records) mapped to GSN (generic sequence number) codes; Free text concatenation of {\tt chief complaint}, {\tt triage assessment} and {\tt MD comments}; age binned by decade; sex; and administrative ICD9 billing codes (used to establish ground truth but not visible during learning or testing).
We apply negation detection to the free-text section using ``negex'' rules \cite{Chapman2001301} with some manual adaptations appropriate for Emergency department notes \cite{JerniteEtAl_nips13health}, and replace common bigrams with a single token (e.g. ``chest pain'' is replaced by ``chest\_pain''.
We reduce the dataset from 273,174 patients to 16,268 by filtering all patients that have fewer than 2 of the manually specified conditions. 
We filter words to remove those that appear in more than $50\%$ of the dataset and take the 1000 most common words after that filtering.
Table~\ref{tab:medical_concepts} lists the concepts that are used and a selection of their anchors specified by a physician research collaborator. 
In the feature vector, anchors are replaced by a single feature which represents a union of the anchors (i.e. whether {\em any } of the anchors appear in the patient record).

{\bf Stack Overflow:} Questions were initially filtered to remove all questions that do not have at least two of the 200 most popular tags. We filter words to remove those that appear in more than $50\%$ of the dataset and take the 1000 most common words after that filtering. Tag names that contain multiple words are treated as N-grams that are replaced by a single token in the text.

\section{Detailed methods}
\subsection{Regularization}
We found it useful to introduce an additional regularization parameter to Equation~\ref{eq:anchor_opt_matrix} that encourages the recovered marginals to be  be close to independent unless there is strong evidence otherwise.
\vspace{-1mm}
\begin{equation}
\label{eq:anchor_opt_matrix-reg}
\mu_{\mathcal{\mathcal{Y}}}^{*} = \argmin_{\mu \in \mathcal{M}} D_{KL} \left(\mu_{\mathcal{A}}, R \mu \right) + \lambda D_{KL} \left(\mu_{indep}, \mu\right),  
\end{equation}
\vspace{-4mm}

Where $\mu_{indep}$ is a marginal vector constructed using the single-variable marginals in an independent distribution. 

For recovery of marginals under local consistency and marginal polytope constraints, we use the the conditional gradient algorithm, as discussed in section~\ref{sec:robust}, using a tolerance of $0.005$ for the duality gap (as described in \citet{jaggi2013revisiting}) as a stopping criterion for Stack Overflow and $0.01$ for medical records.  When using simplex constraints, we use the more appropriate Exponentiated Gradient algorithm~\cite{Kivinen95exponentiatedgradient} for each marginal independently. 
%In the timing experiments, we scale the threshold linearly with the number of recovered marginals.
We use a regularization parameter of $\lambda = 0.01$ to learn $P(\mathcal{Y})$ and $\lambda = 0.1$ to learn $P(\mathcal{X}|\mathcal{Y})$. 
The moments required to learn $P(\mathcal{X}|\mathcal{Y})$ are recovered using only simplex constraints.

Linear programs in the conditional gradient algorithm  are solved using Gurobi \citep{gurobi} and integer linear programs are solved using Toulbar2 \citep{Toulbar2}.
For structure learning we use the gobnilp package \citep{gobnilp} with a BIC score objective, though we note that any exact or approximate structure learner that takes a decomposable score as input could be used equally well.

\subsection{Conditional gradient algorithm for moment recovery}
Algorithm~\ref{alg:FW} describes the conditional gradient~\citep{Frank-Wolfe} algorithm that was use for moment recovery. Line 3 minimizes a linear objective over a compact convex set. In our setting, this is the marginal polytope or its relaxations. To minimize a linear objective over a compact convex set, it suffices to search over the vertices of the set. For the marginal polytope, these correspond to the integral vertices of the local consistency polytope. Thus, this step can be solved as an integer linear program with local consistency constraints. 

We use a ``fully corrective'' variant of the conditional gradient procedure. If the minimization of line 3 returns a vertex that has previously been used, we perform an additional step, moving to the point that minimizes the objective over convex combinations of all previously seen vertices.

\begin{algorithm}[t]
    \caption{Moment recovery (cond. gradient) \label{alg:FW} }
    \mbox{Minimize $f(\mu)=D_{KL}(\mu_A, R\mu)$ s.t. $\mu \in \mathcal{\mathcal{M}}$}
\begin{algorithmic}[1]
    \State initialize $\mu^0 \in \mathcal{\mathcal{M}}$ (e.g. uniform)
    \For{$k = 0,1,\ldots,M$}
    \State \label{algline:FW_min}$s \gets \argmin_{s^{\prime} \in \mathcal{M}}\brangle{s^{\prime},\nabla f(\mu^k)}$
    \State $\text{Compute search direction: }d \gets s - \mu^k$
    \State $\text{Determine stepsize, $\gamma \in [0,1]$}$
    \State $\text{Move in descent direction: }\mu^{k+1} \gets \mu^k + \gamma d^k$
    \EndFor 
\end{algorithmic}
\end{algorithm}

% \subsection{Regularization}
% The regularized objective can be seen as putting a Bayesian prior on the marginal distributions.

% \begin{equation}
% \label{eq:anchor_opt_matrix-reg}
% \mu_{\mathcal{\mathcal{Y}}}^{*} = \argmin_{\mu \in \mathcal{M}} D_{KL} \left(\mu_{\mathcal{A}}, R \mu \right) + \lambda D_{KL} \left(\mu_{indep}, \mu\right),  
% \end{equation}

% We can reframe this as a MAP estimation problem of finding the true marginal vector $\mu$ after observing the marginal vectors of the anchors $\mu_{\mathcal{A}}$ with a Dirichlet prior for $P(\mu)$.

% \begin{gather*}
% \mu_{\mathcal{\mathcal{Y}}}^{*} = \argmax_{\mu \in \mathcal{M}} 
% \log p(\mu_{\mathcal{A}} | \mu) + \log p(\mu)\\
% =\argmax_{\mu \in \mathcal{M}} 
% \sum_i \log (R\mu)^{N \mu_{\mathcal{A}}} + \sum_i \log \mu^{\lambda \mu_{indep}}\\
%  = \argmax_{\mu \in \mathcal{M}} N \sum_i \mu_{\mathcal{A}, i} \log (R\mu)_{i} + \lambda \sum_i \mu_{indep,i} \log \mu_i \\ 
%  = \argmax_{\mu \in \mathcal{M}} N \sum_i \mu_{\mathcal{A}, i} \log (R\mu)_{i} - \sum_i \mu_{\mathcal{A}, i}  \log \mu_{\mathcal{A}, i}   \\
%  + \lambda \sum_i \mu_{indep,i} \log \mu_i  - \sum_i \mu_{indep, i} \log \mu_{indep, i} \\
%   = \argmin_{\mu \in \mathcal{M}} D_{KL} \left(\mu_{\mathcal{A}}, R \mu \right) + \frac{\lambda}{N} D_{KL} \left(\mu_{indep}, \mu\right) 
% \end{gather*}

\subsection{Monte Carlo EM}
When running EM, we optimize the variational lower bound on the data likelihood:
$$\log P(X;\theta) \geq \mathcal{L}(q, \theta) = E_{y \sim q} \left[\log P(x,y; \theta) - \log P(y)\right]$$
The EM algorithm optimizes this bound with a coordinate ascent, alternating between E-steps which improve $q$ and M-steps which improve $\theta$. Usually both the E-step and M-steps are maximization steps, but incomplete M-steps which only improve $\theta$ in every M-step also leads to monotonic improvement of $\log(P(X;\theta)$ with every step. 
In this section, we describe a variant of Monte Carlo EM which is useful for this model using Gibbs sampling to approximate the E-step and a custom M-step which is guaranteed to improve the variational lower bound at every step. We hold the distribution $P(\mathcal{Y})$ fixed and only optimize the failure and leak probabilities. For these purposes, the leak probabilities can be treated as simply failure probabilities of an extra latent variable whose value is always 1.

\subsubsection{Outer E-step}
For the E-step, we use the Gibbs sampling procedure described in 4.3 of \cite{WanSonWan_kdd14}.

\subsubsection{Outer M-step}
\begin{algorithm}[tb]
\caption{Alternative generative model with auxiliary $A$ variables.}
\label{alg:generative_alt}
\begin{algorithmic}
\State $Y \sim P(Y)$
\For{j in 1..m}
\State $A_j \sim P(A_j = k | Y)$
\State $X_j = A_j \leq n$
\EndFor
\end{algorithmic}
$P(A_j = k | Y) = \begin{cases} (1-f_{k,j})\prod_{i=0}^{k-1} f_{i,j}^{y_i} & k \leq n\\
                               1- \prod_{i=0}^{k-1} f_{i,j}^{y_i} & k = n+1
                 \end{cases}$
\end{algorithm} 

The M-step consists of a coordinate step guaranteed to improve $\theta$ for a fixed $q$. 
$P(x,y ; \theta)$ is a fully observed model, but optimizing $\theta$ has no closed form. Instead we introduce $m$ {\em auxiliary variables} $A \in [0,n+1]^m$, and adopt the generative model described in Algorithm~\ref{alg:generative_alt} which equivalently describes the fully-observed noisy-or model (i.e. $\sum_a P(X,Y,a;\theta) = P_{\text{noisy or}}(X,Y;\theta)$).
In this new expanded model, we can perform a single E-M step with respect to the latent variable $A$ in closed form.

{\bf inner E-step:}
\begin{equation}
P(A_j = k|X_j,Y) = 
\begin{cases}
1 & X_j = 0 \cap k = n+1 \\
0 & X_j = 0 \cap k \neq n+1 \\
0 & X_j = 1 \cap k = n+1 \\
0 & X_j = 1 \cap k \leq n \cap Y_k = 0 \\
\propto \prod_{i=0}^{k-1}f_{i,j}^{y_i}(1-f_{k,j}) & X_j = 1 \cap k \leq n \cap Y_k = 1 \\
\end{cases}
\end{equation}

{\bf inner M-step:}
\begin{equation}
f_{i,j} = 1-\frac{\text{count}(A_j = i)}{\text{count}(A_j \leq i)}
\end{equation}

\section{Learned models}
\subsection{Tree models}
\begin{figure*}[ht]
\centering
\includegraphics[scale=0.5, angle=90]{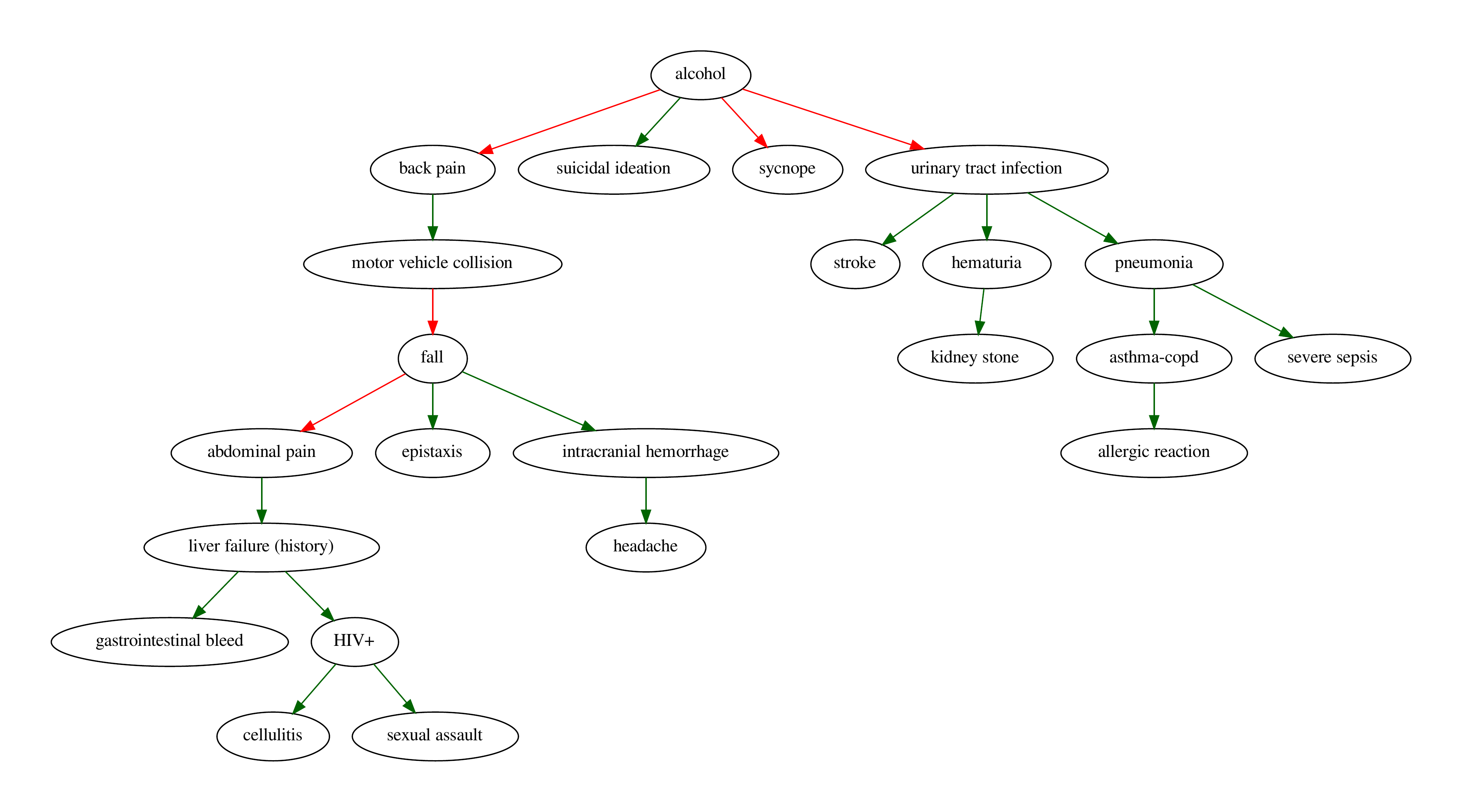}
\caption{\label{fig:tree-emergency} Tree model learned for Emergency data. Red and green edges represent positive and negative correlations between the variables, respectively.}
\end{figure*}

\begin{figure*}[ht]
\centering
\includegraphics[scale=0.35,angle=90]{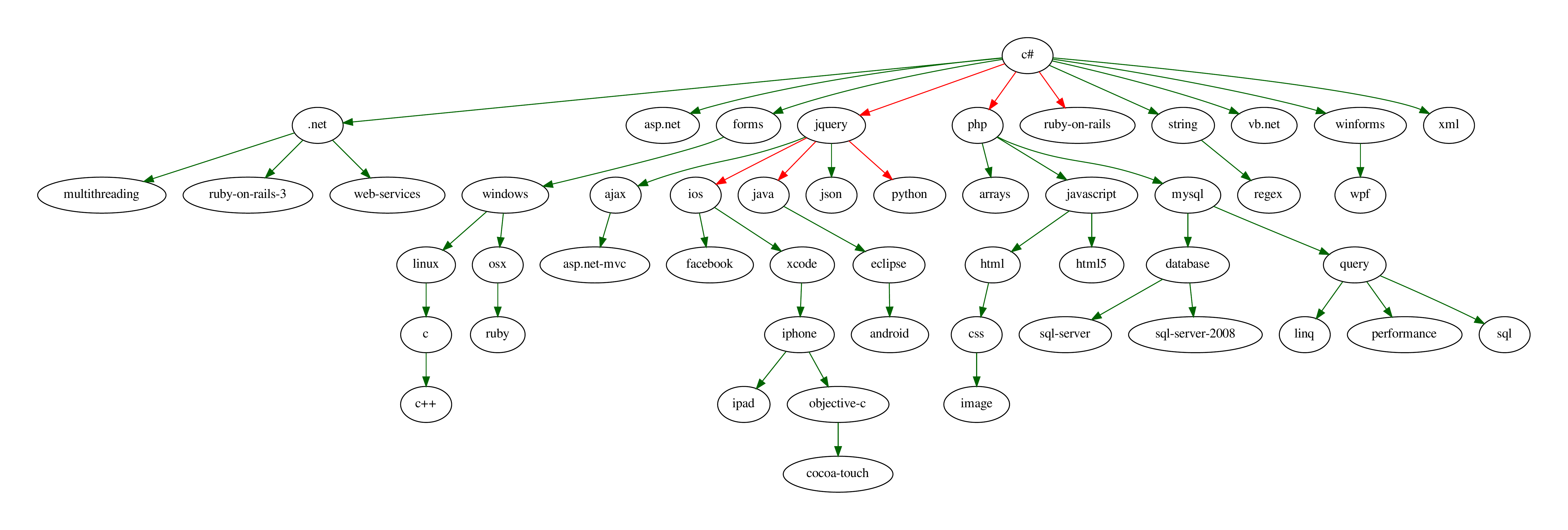}
\caption{\label{fig:tree-stackoverflow} Tree model learned for Stack Overflow data. Red and green edges represent positive and negative correlations between the variables, respectively.}
\end{figure*}

Figures~\ref{fig:tree-emergency} and \ref{fig:tree-stackoverflow} show models learned using second order recovered moments to learn maximal scoring tree structured models. Tables~\ref{tab:medical_concepts} and \ref{tab:programming_concepts} show the factor loadings learned for these tree-structured models.


\subsection{Bounded in-degree models}
\begin{figure}[ht]
\centering
\includegraphics[scale=0.5]{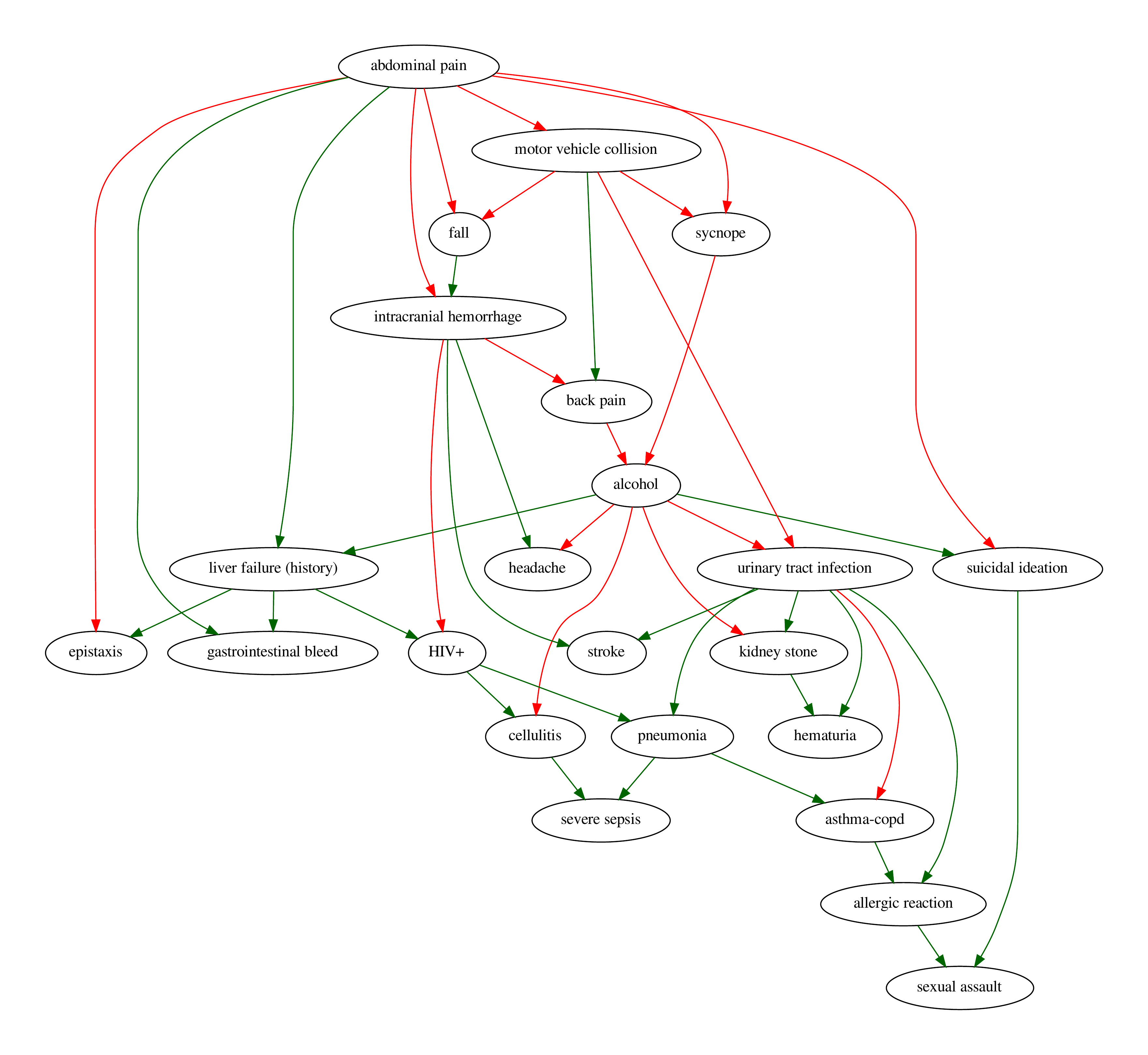}
\caption{\label{fig:graph-emergency} Bounded in-degree model ($\leq 2$)learned for Emergency data. Red and green edges represent positive and negative correlations between the variables, respectively. }
\end{figure}

\begin{figure}[ht]
\centering
\includegraphics[scale=0.35]{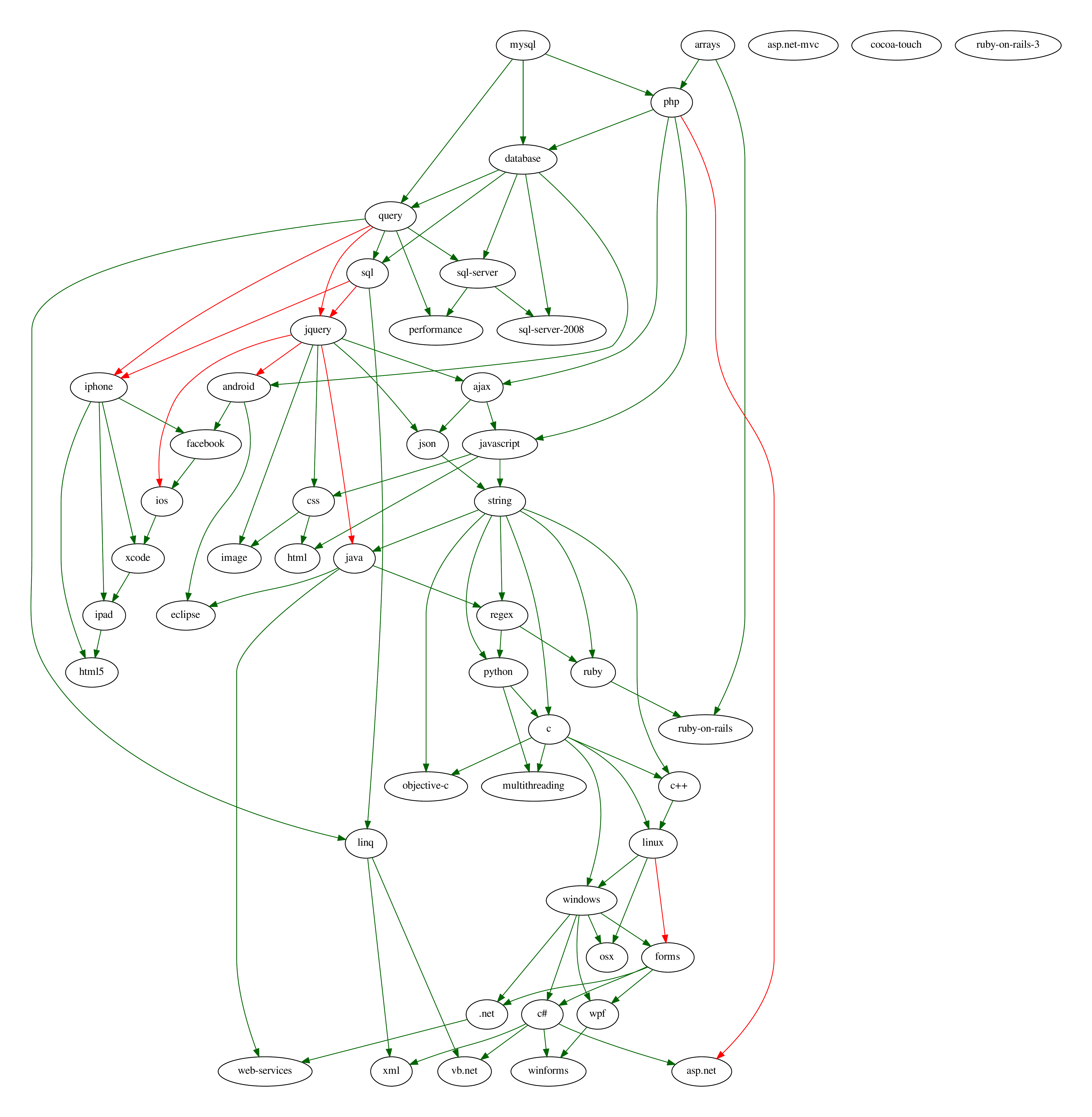}
\caption{\label{fig:graph-stackoverflow} Bounded in-degree model ($\leq 2$) learned for Stack Overflow data. Red and green edges represent positive and negative correlations between the variables, respectively. }
\end{figure}

Figures~\ref{fig:graph-emergency} and \ref{fig:graph-stackoverflow} show models learned using third order recovered moments to learn maximal scoring graphs with bounded in-degree of two.

\afterpage{%
\begin{table}[tb]
\small
\caption{
    \label{tab:medical_concepts}
    Learned medical concepts. For each concept we display the top 10 weighted factors and one supplied anchor.}
\begin{tabular}{|l|p{10cm}|l|}
\hline
 \rowcolor{lightgray}
{\bf Latent variable name} & {\bf Top weights} &{\bf anchors} \\
\hline
\input{tex/emergency_factors}
\hline
\end{tabular}
\end{table}

\begin{longtable}{|l|p{10cm}|l|}
\caption{Learned concepts from Stack Overflow. For each concept we display the top weighted factors and one supplied anchor. Note that in the Stack Overflow dataset we use a simple rule to provide a single anchor for every latent variable.\label{tab:programming_concepts}}\\
\hline
 \rowcolor{lightgray}
{\bf Latent variable name} & {\bf Top weights} &{\bf anchors}\\
\hline
osx & osx, i'm, running, i've, install, installed, os, code, managed, existing & osx \\
ruby-on-rails-3 & parsing, executed, web-application, don't, numbers, developed, resources, dynamic, named, rest & ruby-on-rails-3 \\
image & image, code, size, upload, html, save, picture, width, page, display & image \\
mysql & mysql, query, rows, row, id, 1, tables, join, insert, column & mysql \\
web-services & web, service, web-services, client, java, services, things, sharepoint, don't, true & web-services \\
objective-c & objective-c, i'm, code, don't, xcode, dynamic, syntax, including, follow, trouble & objective-c \\
linux & linux, running, command, machine, system, server, directory, install, php, servers & linux \\
query & query, table, a, result, results, queries, tables, return, returns, select & query \\
regex & regex, match, expression, regular, pattern, i'm, replace, characters, html, extract & regex \\
php & php, server, array, send, data, i'm, database, output, website, user & php \\
java & java, string, program, client, xml, read, code, existing, send, environment & java \\
ruby-on-rails & ruby-on-rails, rails, i'm, database, user, controller, page, ruby, model, code & ruby-on-rails \\
asp.net & asp.net, asp.net-mvc-3, controller, site, website, database, control, ajax, jquery, action & asp.net \\
sql-server & sql-server, query, tsql, sql, sql-server-2005, server, stored, procedure, rows, columns & sql-server \\
forms & form, forms, page, code, data, html, jquery, user, button, submit & forms \\
python & python, i'm, a, program, string, list, module, output, don't, write & python \\
json & json, string, response, data, php, parse, code, format, ajax, javascript & json \\
html & html, tags, tag, page, website, code, links, show, webpage, text & html \\
iphone & iphone, app, view, device, sqlite, api, video, uitableview, images, ios5 & iphone \\
performance & performance, time, data, question, takes, slow, faster, run, seconds, running & performance \\
android & android, code, xml, java, activity, device, app, phone, screen, android-layout & android \\
multithreading & thread, multithreading, code, threads, application, method, data, main, run, managed & multithreading \\
xcode & xcode, project, build, app, 4, version, i've, running, debug, target & xcode \\
css & css, page, css3, jquery, style, javascript, chrome, works, width, browser & css \\
jquery & jquery, html, works, plugin, javascript, jquery-ui, css, jquery-ajax, elements, link & jquery \\
string & string, a, strings, array, characters, output, convert, json, character, split & string \\
c\# & c\#, function, code, methods, event, excel, click, written, program, call & c\ \\
javascript & javascript, jquery, js, script, works, ajax, php, button, browser, code & javascript \\
ios & ios, iphone, app, ios5, device, i'm, ipad, 5, user, screen & ios \\
linq & linq, linq-to-sql, code, class, c\#, error, xml, expression, property, collection & linq \\
xml & xml, parse, document, read, data, string, node, output, format, tag & xml \\
ajax & ajax, code, javascript, jquery-ajax, works, call, response, user, request, html & ajax \\
facebook & facebook, users, login, api, app, sdk, post, php, id, link & facebook \\
html5 & html5, html, browser, chrome, css, support, image, browsers, works, android & html5 \\
sql & sql, table, database, query, data, tsql, tables, statement, column, sql-server & sql \\
wpf & wpf, control, window, property, binding, a, bind, controls, ui, items & wpf \\
asp.net-mvc & code, existing, browsers, query, faster, issues, row, flash, environment, program & asp.net-mvc \\
ruby & ruby, i'm, rails, running, string, install, installed, array, feed, executed & ruby \\
ipad & ipad, works, fine, screen, ios, page, device, problem, safari, app & ipad \\
c & c, program, a, write, c+, library, compile, string, array, functions & c \\
database & database, a, table, user, db, code, sql, data, store, created & database \\
arrays & arrays, i'm, 2, data, dynamic, results, saved, row, 0, finally & arrays \\
vb.net & vb.net, string, project, work, database, code, developed, results, rest, existing & vb.net \\
.net & .net, framework, i've, windows, executed, managed, valid, don't, developed, existing & .net \\
eclipse & eclipse, project, build, installed, plugin, debug, running, folder, version, projects & eclipse \\
c++ & c++, c+, code, i'm, managed, find, access, application, implementation, writing & c++ \\
windows & windows, windows-7, a, running, winapi, c\#, machine, run, application, window & windows \\
winforms & winforms, control, user, i'm, windows, a, .net, controls, forms, don't & winforms \\
cocoa-touch & a, i'm, cocoa-touch, code, dynamic, results, follow, row, send, environment & cocoa-touch \\
sql-server-2008 & sql-server-2008, sql-server, sql, query, tsql, stored, 1, procedure, developed, resources & sql-server-2008 \\
\hline
\end{longtable}
}

%% file: tex/emergency_factors.tex
%auto-ignore
abdominal pain & pain, dispensed:ondansetron, nausea, neg:fevers, dispensed:morphine sulfate, vomiting, days, dispensed:hydromorphone (dilaudid), neg:vomiting & abdominal pain \\
alcohol & sex:m, sober, admits, found, drink, dispensed:thiamine, dispensed:folic acid, dispensed:diazepam, dispensed:multivitamins, dispensed:multivitamins & etoh \\
allergic reaction & disposition, initial vitals, pending, consults, interventions, trigger, imaging, ed, diagnosis, pain & allergy \\
asthma-copd & med-history:albuterol sulfate, sob, dispensed:methylprednisolone sodium succ, cough, nebs, med-history:spiriva with handihaler, home, days, dyspnea & asthma \\
back pain & pain, neg:or, neg:pain, back, denies, lower back, dispensed:oxycodone-acetaminophen, neg:of, low back, neg:bowel & back pain \\
cellulitis & ed, admit, swelling, imaging, diagnosis, days, consults, iv, leg, interventions & cellulitis \\
stroke & age 80, admit, patient, head, ekg, weakness, ed, ct head, neuro, htn & stroke \\
epistaxis & nose, bleeding, blood, neg:bleeding, ekg, neg:with, ed, bleed, consults, today & epistaxis \\
fall & pain, denies, neg:pain, p fall, neg:or, ed, imaging, consults, interventions, neg:loc & fell down \\
gastrointestinal bleed & blood, dispensed:pantoprazole sodium, hct, gi, stool, admit, rectal, today, dispensed:lidocaine jelly 2\% (urojet), & gi bleed \\
headache & head, nausea, today, neg:or, denies, neg:pain, days, dispensed:acetaminophen, pain, neck pain & headache \\
hematuria & sex:m, urine, urology, blood, foley, neg:pain, days, neg:fevers, dysuria, bladder & hematuria \\
hiv+ & sex:m, med-history:truvada, cd, age 40, med-history:ritonavir, days, pcp, cough, fever, denies & hiv \\
intracranial hemorrhage & osh, ct, fall, age 80, found, head ct, transfer, sex:m, repeat, small & ich \\
kidney stone & pain, flank pain, dispensed:ondansetron, dispensed:ketorolac, stone, nausea, urology, ct, dispensed:morphine sulfate & kidney stone \\
liver failure (history) & sex:m, liver, age 50, med-history:lactulose, admit, ed, med-history:folic acid, med-history:furosemide, med-history:multivitamin, med-history:lasix & cirrhosis \\
motor vehicle collision & car, neg:loc, age 20, hit, neck, neg:head, driver, mph, neg:airbag, front & mvc \\
pneumonia & cxr, admit, sex:m, cough, ed, fever, diagnosis, ekg, med-historyicine, admit med-historyicine & pna \\
severe sepsis & dispensed:fentanyl citrate, found, icu, osh, vanc, age 80, imaging, consults, interventions, lactate & severe sepsis \\
sexual assault & patient, dispensed:ondansetron odt, evaluation, ceftriaxone, man, eval, age 30, home, plan, flagyl & sane nurse \\
suicidal ideation & depression, neg:hi, denies, states, neg:si, plan, dispensed:lorazepam, eval, age 40, section & si \\
sycnope & ekg, neg:pain, fall, head, fell, denies, neg:cp, ed, neg:of, loc & syncope \\
urinary tract infection & ed, imaging, consults, interventions, diagnosis, ua, cipro, admit, pending, disposition & uti \\